\documentclass[conference]{IEEEtran}
\IEEEoverridecommandlockouts
\usepackage{cite}
\usepackage{amsmath,amssymb,amsfonts}
\usepackage{algorithmic}
\usepackage{graphicx}
\usepackage{textcomp}
\usepackage{xcolor}
\usepackage{caption}
\def\BibTeX{{\rm B\kern-.05em{\sc i\kern-.025em b}\kern-.08em
 T\kern-.1667em\lower.7ex\hbox{E}\kern-.125emX}}
\begin{document}

\title{Temporal Spatial Decomposition and Fusion Network for Time Series Forecasting*\\

}

\author{\IEEEauthorblockN{1\textsuperscript{st} Liwang Zhou}
\IEEEauthorblockA{\textit{Zhejiang University} \\
China \\
21731005@zju.edu.cn}
\and
\IEEEauthorblockN{2\textsuperscript{nd} Jing Gao}
\IEEEauthorblockA{\textit{Anhui University} \\
China \\
jingles980@gmail.com }
}

\maketitle

\begin{abstract}
Feature engineering is required to obtain better results for time series forecasting, and decomposition is a crucial one. One decomposition approach often cannot be used for numerous forecasting tasks since the standard time series decomposition lacks flexibility and robustness. Traditional feature selection relies heavily on preexisting domain knowledge, has no generic methodology, and requires a lot of labor. However, most time series prediction models based on deep learning typically suffer from interpretability issue, so the "black box" results lead to a lack of confidence. To deal with the above issues forms the motivation of the thesis. In the paper we propose TSDFNet as a neural network with self-decomposition mechanism and an attentive feature fusion mechanism, It abandons feature engineering as a preprocessing convention and creatively integrates it as an internal module with the deep model. The self-decomposition mechanism empowers TSDFNet with extensible and adaptive decomposition capabilities for any time series, users can choose their own basis functions to decompose the sequence into temporal and generalized spatial dimensions. Attentive feature fusion mechanism has the ability to capture the importance of external variables and the causality with target variables. It can automatically suppress the unimportant features while enhancing the effective ones, so that users do not have to struggle with feature selection. Moreover, TSDFNet is easy to look into the "black box" of the deep neural network by feature visualization and analyze the prediction results. We demonstrate performance improvements over existing widely accepted models on more than a dozen datasets, and three experiments showcase the interpretability of TSDFNet.
\end{abstract}


\section{Introduction}
Time series forecasting plays a key role in numerous fields such as economy\cite{b1},finance\cite{b2}, transportation\cite{b3}, meteorology\cite{b4}, It empowers people to foresee opportunities and serves as guidance for decision-making. Therefore, it is crucial to increase the generality of time series models and lower modeling complexity while maintaining performance. In the field of time series forecasting, multi-variable and multi-step forecasting forms one of the most challenging tasks. Errors may accumulate as the forecast step increases.At present, there is no universal method to handle the problem of multi-variable and multi-step time series prediction. One time series usually calls for its specific feature engineering and forecasting model, due to the complexity and diversity of real world time series, which usually requires data analysts to have specialized background knowledge.

Feature engineering is usually used to preprocess data before modeling. In the field of feature engineering, time series decomposition is a classical method to decompose a complex time series into numerous predictable sub-series, such as STL \cite{b48} with seasonal and trend decomposition, EEMD\cite{b27} with ensemble empirical mode decomposition, EWT\cite{b28} with empirical wavelet transform. In addition, feature selection is another important step. For complex tasks, some auxiliary variables are usually needed to assist the prediction of target variables. The reasonable selection of additional features is crucial to the performance of the model, because the introduction of some redundant additional features may degrade the performance of the model. How to choose the appropriate decomposition methods and important additional features is also a challenging problem for data analysts.

On the other hand, despite the fact that numerous models have been put forth, each one has drawbacks of its own. The majority of deep learning based models are difficult to comprehend and produce unconvincing predictions. However, models like ARIMA and xgboost\cite{b29}, which have sound mathematical foundations and offer interpretability, cannot compete with deep learning-based models in terms of performance.

Therefore, it is necessary to break the traditional practice and devise a new way to handle these problems.
In this study, a novel neural network model called TSDFNet is developed based on the self-decomposition mechanism and attentive feature fusing mechanism. Decomposition and feature selection are integrated as internal modules of the deep model to lessen complexity and increase adaptability. The data's high-order statistical features may be captured by this model's robust feature expression capabilities, which make it applicable to datasets from a variety of domains.
 
 In summary, The contributions are summarized as follows:
 \begin{itemize}
 \item We proposed Temporal Decomposition Network (TDN), which is extensible and adaptive.it decomposes time series over temporal dimension and allows users to customize basis functions for specific tasks.
 \item We proposed Spatial Decomposition Network (SDN), which creatively uses high-dimensional external features as decomposition basis functions to model the relationship between external variables and target variables.
 \item We proposed Attentive Feature Fusion Network (AFFN), which has the ability of automatic feature selection and can capture the importance and causality of features. In this way, users can avoid the trouble of feature selection and use arbitrary basis functions in the self-decomposition network without worrying about the loss of model performance caused by introducing invalid features.
 \item TSDFNet obtains interpretable results on datasets in multiple fields, and has significantly improved performance compared with many traditional models.
\end{itemize}

\section{Relative Work}

The field of time series prediction has a rich history, and many outstanding models have been developed. The most well-known conventional methods include ARIMA\cite{b8} and exponential smoothing\cite{b9}. The interpretability and usability of the ARIMA model, which turns nonstationary processes into stationary ones through difference and can also be further expanded into VAR\cite{b10} to address the issue of multivariate time series forecasting, are the main reasons for its popularity. Another effective forecasting technique is exponential smoothing, which smooths univariate time-series by giving the data weights that decrease exponentially over time.

Since time series prediction is essentially a regression problem, it is also possible to utilize a variety of regression models.Some machine learning-based techniques, including decision trees \cite{b13} and support vector regression (SVR) \cite{b12}.
Additionally, ensemble methods, which employ multiple learning algorithms to achieve better predictive performance than could be attained from any one of the constituent learning algorithms alone, are effective tools for sequence prediction. Examples of these methods include random forest \cite{b14} and adaptive lifting algorithm (Adaboost)\cite{b15}.

In recent years, deep learning has become popular and neural networks have achieved success in many fields \cite{b32}, \cite{b33}, \cite{b34}. It uses the back propagation algorithm \cite{b35} to optimize the network parameters.
Long Short-Term Memory (LSTM)\cite{b16} and its derivatives shows great power in sequential data, It overcomes the defect of vanishing gradient of recurrent neural network (RNN) \cite{b17} and can better capture long-term dependence.
Deep autoregressive network (DeepAR) \cite{b18} uses stacked LSTMs for iterative multi-step prediction, and Deep state-space Models (DSSM) \cite{b19} also adopts a similar approach, utilizing LSTMs to generate parameters of a predefined linear state-space model.
Sequence to Sequence (Seq2Seq) \cite{b20} usually uses a pair of LSTMs or GRUs \cite{b21} as encoder and decoder. The encoder maps the input data to the hidden space into a fixed-length semantic vector, The decoder reads the context vector and attempts to predict the target variable step by step. Temporal convolutional network (TCN) \cite{b22} could also be effectively applied to the sequence prediction problem, which may be used as an alternative to the popular RNN family of methods and has faster speed and fewer parameters compared with RNN-based models with causal convolution and residual connection.
The attention mechanism \cite{b25} emerged as an improvement over the encoder decoder based \cite{b26}, and it can easily be further extended into a self-attentional mechanism as the core of the Transfomer models\cite{b23}, \cite{b24}.

\section{Methodology}
\begin{figure}[htbp] 
 \centering
 \includegraphics[width=\linewidth]{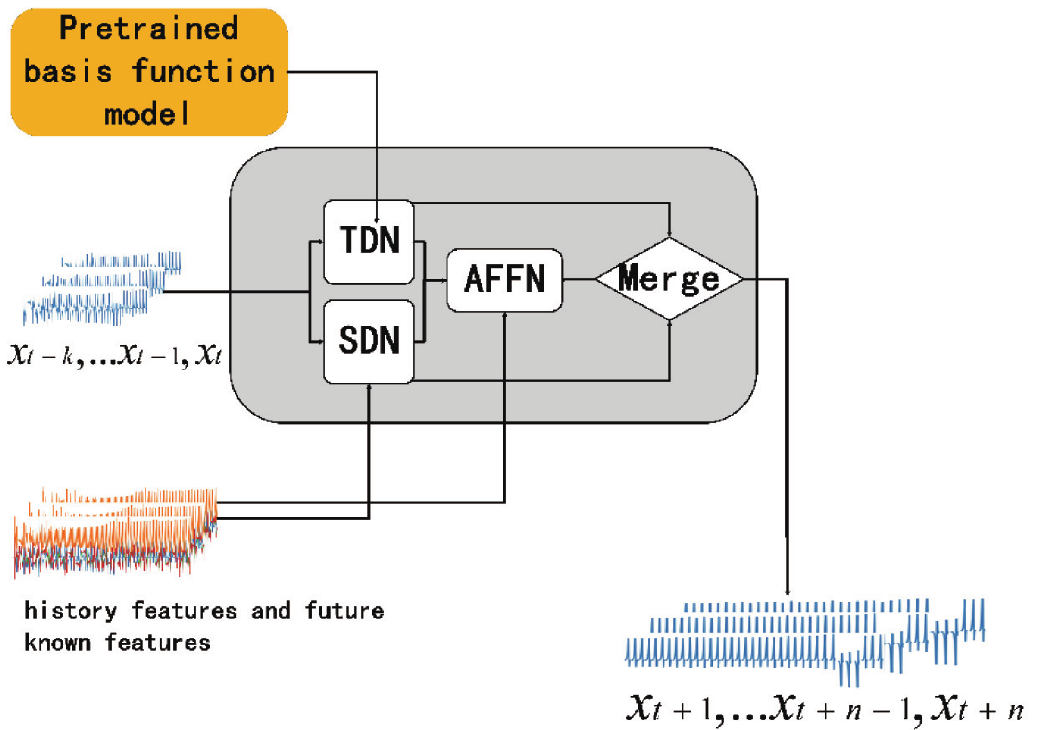}
 \caption{Overall structure of TSDFNet}
 \label{fig:3-1}
\end{figure}

The network's architecture is depicted in Figure \ref{fig:3-1}. It has two main parts, the first of which is a self-decomposition network that includes TDN and SDN. The feature fusion network (AFFN), is the additional element.

\subsection{Self-decomposing network }\label{AA}

The structure of self-decomposition network includes two decomposition modules, one is time decomposition network TDN, which adopts custom basis function to decompose sequences in time dimension. The other is spatial decomposition network SDN, which decomposes sequences in generalized spatial dimensions, using exogenous features as basic functions. Its main objective is to break down complex sequences into ones that are simple and predictable.

TDN uses multiple sets of pre-trained basis functions with different parameters to capture signal features, which could be triangular basis, polynomial basis, wavelet basis and so on.

 \begin{figure}[htbp]
 \centering
 \includegraphics[width=\linewidth]{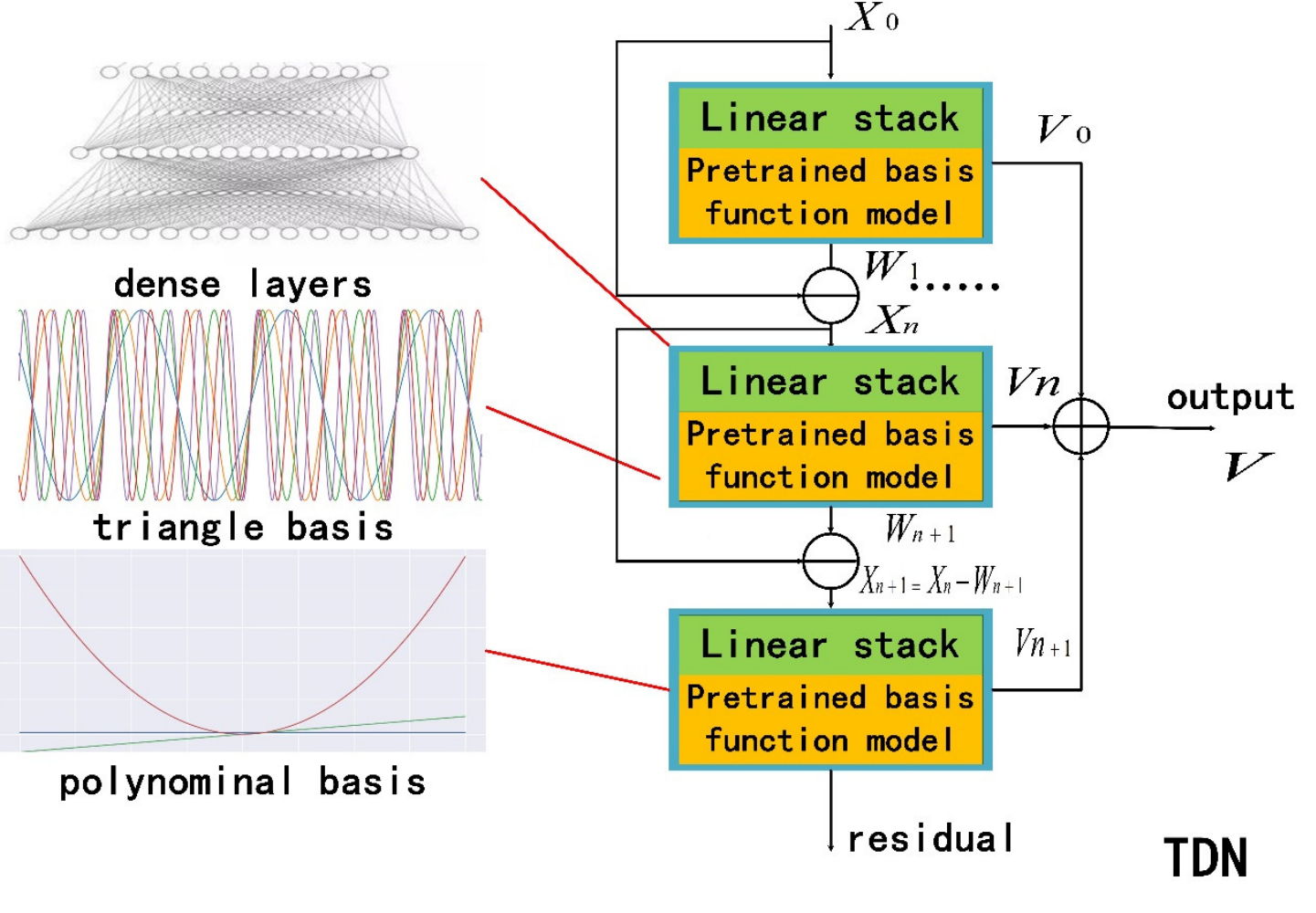}
 \caption{Temporal Decomposition Network}
 \label{fig:3-2}
 \end{figure}

The architecture of TDN is shown in Figure \ref{fig:3-2}. There are $N$ recursive decomposition units in it. $(n+1)^{th}$ unit accepts its respective input $\mathbf{X}_{n}$ as input and output two intermediate components $\mathbf{W}_n$ and $\mathbf{V}_n$. Each decomposition unit consists of two parts, Stacked fully connected network $L_s$ maps data into hidden space to produce the semantic vector $\mathbf{S}_{n}$, predicts basis expansion coefficients both forward and backward through two sets of fully connected networks $L_p$ and $L_q$ respectively. The process is:

\begin{flalign}
 && \mathbf{S}_{n} & = L_s(\mathbf{X}_{n}) & 
\end{flalign}
\begin{flalign}
 && \mathbf{P}_{n} & = L_p(\mathbf{S}_{n}) & 
\end{flalign}
\begin{flalign}
 && \mathbf{Q}_{n} & = L_q(\mathbf{S}_{n}) &
\end{flalign}
The other part is a group of pretrained basis function models, which are functions of the time vector $\mathbf{t}=[-w,-w+1,...0,h-1,h]/L$ defined on a linear space from $-w/L$ to $h/L$, where $L=w+h+1$,$w$ is the time historical window length of drive sequence, $h$ is the time window length of the target sequence. This time vector is fed into different pre-trained models and mapped into multiple basis functions, such as trigonometric functions with different frequencies defined by $\mathbf{C}_{n}=[sin(-k\mathbf{t}),cos(-k\mathbf{t}),...,cos(k\mathbf{t}),sin(k\mathbf{t})]$, polynomial function with different degrees defined by $\mathbf{C}_{n}=[\mathbf{t},\mathbf{t}^2,\mathbf{t}^3...\mathbf{t}^k]$.
$\mathbf{C}_{n}$ is divided into $\mathbf{C}^p_{n}$ when $\mathbf{t} = [-w/L...0)$ and $\mathbf{C}^p_{n}$ when $\mathbf{t} = [0...h/L]$, which are used to fit historical and future data respectively. Their coefficient matrix $\mathbf{P}_{n}$ and $\mathbf{Q}_{n}$ determine the importance of each basis function. The final outputs of $n^{th}$ block are defined by:
\begin{flalign}
 && \mathbf{W}_{n} & = \mathbf{C}^p_{n}\mathbf{P}_{n} & 
 \label{eq:wn}
\end{flalign}
\begin{flalign}
 && \mathbf{V}_{n} & = \mathbf{C}^q_{n}\mathbf{Q}_{n} & 
 \label{eq:vn}
\end{flalign}
The input to the next block is defined as follows:
\begin{flalign}
 && \mathbf{X}_{n + 1} & = \mathbf{X}_{n} - \mathbf{W}_{n} & \label{eq:1}
\end{flalign}
The original signal $X_0=[x_{t-w},x_{t-w+1}...,x_{t - 1}]$ keeps removing backward feature components $W_n$, and ideally the residual is a random noise that no longer contains feature information. The process is:
\begin{flalign}
 && residual & = \mathbf{X}_{0} - \sum\limits_{i = 0}^{n + 1} {\mathbf{W}_{i}} & 
\end{flalign}
The forward output of each decomposition unit is accumulated to form the TDN output as follow:
\begin{flalign}
 && \mathbf{V} & = \sum\limits_{i = 0}^{n + 1} {\mathbf{V}_{i}} &
\end{flalign}

The model's hyperparameter $N$, which is defined as the number of decomposition layers, is dependent on the kind of basis function you select and is repeatable for each basis function. The weights of these basis function networks can be adjusted once again for different tasks to adapt to diverse scenarios.This means that $\mathbf{C_n}$ is not fixed, it will be independent learning into other comparable forms. 
\begin{figure}[htbp]
 \centering
 \includegraphics[width=\linewidth]{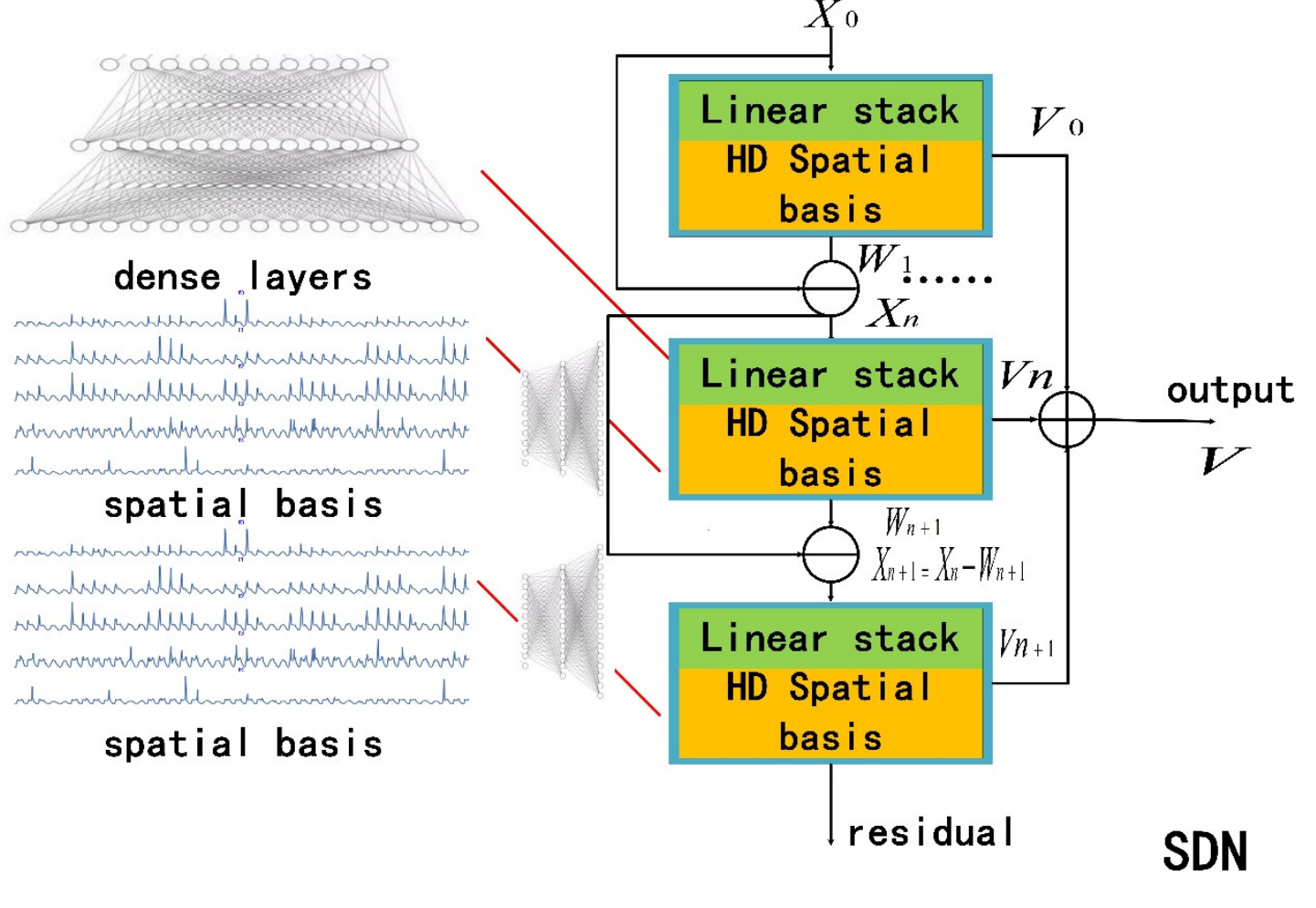}
 \caption{Spatial Decomposition Network}
 \label{fig:3-3}
\end{figure}

 Spatial decomposition network (SDN) is shown in Figure \ref{fig:3-3}. Its structure is similar to TDN. The difference is that SDN adopts external features mapped to higher dimensions as the basis vectors $\mathbf{C}^p_{n}$ and $\mathbf{C}^q_{n}$, Details are shown as follows:
\begin{flalign}
 && \mathbf{C}^p_{n} & = L_{n}(\mathbf{E}_p) &
 \label{eq:9}
 %(3.9)
\end{flalign}
\begin{flalign}
 && \mathbf{C}^q_{n} & = L_{n}(\mathbf{E}_q) &
 \label{eq:10}
 %(3.10)
\end{flalign}
where $\mathbf{E}_p$ is the historical additional feature and $\mathbf{E}_q$ is the future additional feature. $L_n$ are stacked fully connected modules that map historical and future additional features to higher dimensions, respectively.
We employ embedding to transform some discrete features, and use 0 to fill in the missing features.Additionally, it is important to note that the self-decomposition module offers flexibility to handle the univariable time series. The module of the spatial decomposition network (SDN) can be disabled, or alternatively, SDN can be enabled and accept feature components of other methods as input, such as EMD \cite{b45}.

\begin{figure}[htbp]
 \centering
 \includegraphics[width=\linewidth]{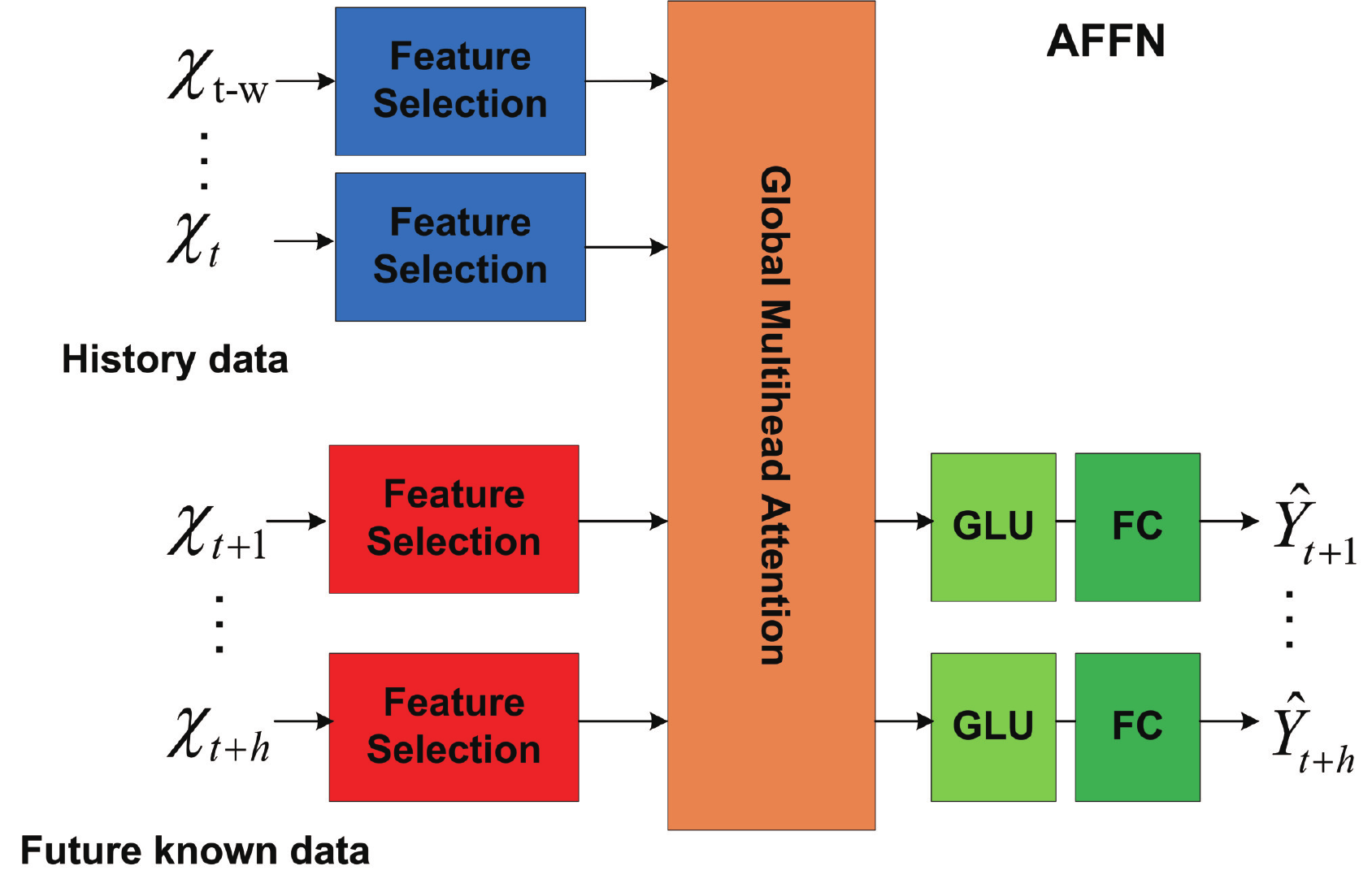}
 \caption{Attentive Feature Fusion Network}
 \label{fig:3-4}
\end{figure}
\subsection{Attentive Feature Fusion network }

Attentive feature fusion network as shown in Figure \ref{fig:3-4}, the feature selection module is designed to provide instance-wise variable selection, a groups of decision units iteratively suppress irrelevant features, the multihead attention block accepts the result of the feature selection modules as input to further model global relationships. Finally, the output is obtained by the GLU \cite{b56} followed by FC at each time step, Here GLU attemps to suppress the invalid part of the input data.

\begin{figure}[htbp]
 \centering
 \includegraphics[width=\linewidth]{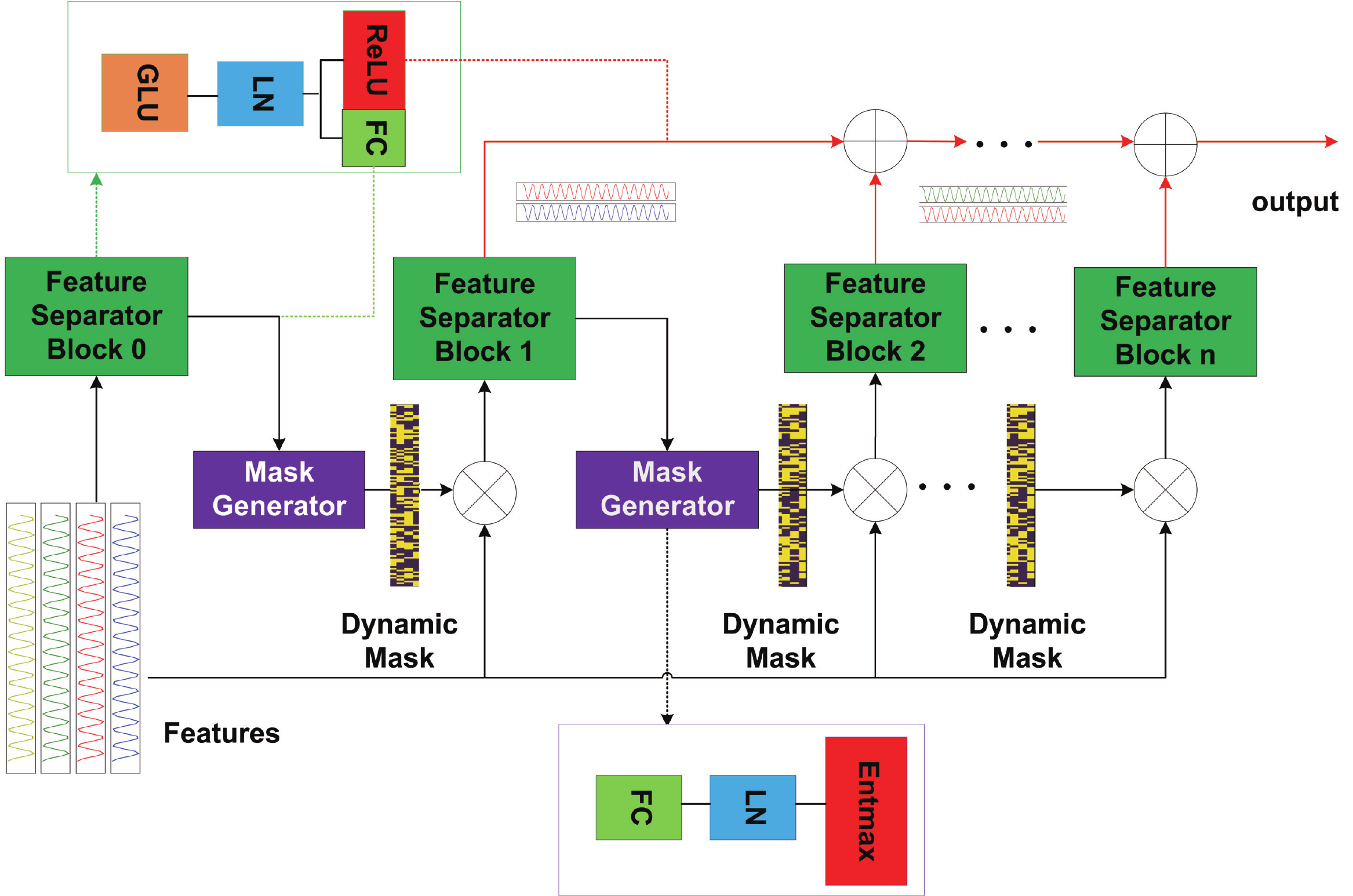}
 \caption{Feature Selection Network}
 \label{fig:3-5}
\end{figure}

We designed a feature selection block as illustrated in Figure \ref{fig:3-5}, including the mask generator and the feature separator.

The learnable mask generator is employed to produce a dynamic mask $\mathbf{M}$, which has explain-ability for indicating the importance of each dimension feature. The Mask at decision step $j$ takes the separated feature from step $j-1$ as input. $\mathbf{M}_j=entmax(h(a_{j-1})$, $h$ is a trainable function, $entmax$\cite{b30} is similar to the traditional softmax, which has the ability to map vectors to sparse probabilities. a series of basis function feature could be filtered by $\mathbf{M}$, if $\mathbf{M}_{j}$=0, it means that feature input in this step is irrelevant for the prediction task as this sample, the filtered features $\mathbf{M}\cdot \mathbf{X} $ are provided to the next decision step. Generally, we are interested in which feature has more contribution to the target variable. Assuming that the dataset has batch size $B$, the time length is $T$, and the decision step is $J$, the importance distribution could be obtained from

\begin{flalign}
 && \mathbf{D} & = \frac{1}{B*T*J} \sum\limits_{b = 1}^{B} \sum\limits_{t = 0}^{T} \sum\limits_{j = 1}^{J} {\mathbf{M}_{b,t,j}}&
 \label{eq:dist}
\end{flalign}

Feature separator has shared layers, which is composed of a GLU, LN\cite{b31}. Layer normalization (LN) is a technique to normalize the distributions of intermediate layers. It enables smoother gradients,faster training, and better generalization accuracy. the output of share layers $\mathbf{s}$ in feature separator is divide into two parts $[\mathbf{s_1},\mathbf{s_2}]$, one of them $f_1(\mathbf{s_1})$ are used as input of mask generator,another $f_2(\mathbf{s_2})$ is as output of $j^{th}$ level of decision, where $f_1$ is FC layer, $f_2$ is $Relu$, if the element in $\mathbf{s_2}<0$, it has no contribution to current decision output.

Multiple attention was initially applied to language models,and details are shown as follows:
\begin{flalign}
 && A({\bf{Q}},{\bf{K}}) & = Softmax ({\bf{Q}}{{\bf{K}}^T}/\sqrt {d_{attn}} ) &
 \label{eq:softmax}
\end{flalign}
\begin{flalign}
 && Attention({\bf{Q,K,V}}) & = A({\bf{Q,K}}){\bf{V}} &
 \label{eq:attention}
\end{flalign}
\begin{flalign}
 && {\bf{head}}_{i} = & 
 Attention({\bf{QW}}{{\bf_{i}}^Q},{\bf{KW}}{{\bf_{i}}^K},{\bf{VW}}{{\bf_{i}}^V})& 
 \label{eq:head}
\end{flalign}
\begin{flalign}
 && MultiHead({\bf{Q}},{\bf{K}},{\bf{V}}) = & 
 Concat({\bf{head_{1}}},...,{\bf{head_{h}}}){{\bf{W}}^O} &
 \label{eq:multihead}
\end{flalign}
Among them,$MultiHead({\bf{Q}},{\bf{K}},{\bf{V}})$ represents the Attention function.$Softmax$ is the probability distribution function and a parameter $d_{attn}$ used to normalize features on a scale; $\mathbf{K}$ is the key of a certain time segment and $\mathbf{V}$ is value of features. $Q$ is the query feature of the input, ${\mathbf{W}^O}$ is the weight of the network output. The model uses the self-attentional mechanism to learn the correlation between each feature at various times. For the query of an element in a given target, the weight coefficient of each key to the target is obtained by calculating the similarity between query $\mathbf{Q}$ and $\mathbf{K}$, and then the weighted sum of the target is carried out to obtain the final attention value. By dividing the model into multiple heads and forming different subspaces, the model can focus on different aspects of the feature. In this model, $\mathbf{Q}$ is the high-dimensional feature of all historical information  output from the feature selection module, while $\mathbf{K}$ and $\mathbf{V}$ are the high-dimensional feature of future known information.

\section{Experiment}
To evaluate the proposed TSDFNet, we chose a number of typical models, including TCN based on causal convolution, Lstnet \cite{b46} based on CNN and RNN, Seq2Seq based on attention and LSTM,Lstm-SAE \cite{b47} based on stacked autoencoder, for thorough comparison on various types of datasets .

\subsection{Implementation details} 

The experiment is based on Pytorch framework in Ubuntu system. Hardware configuration is Intel(R) Core(TM) I7-6800K CPU @ 3.40 GHZ, 64 GB memory, GeForce GTX 1080 GPU.

First, we construct several sets of synthetic basis functions with different parameters $k$, which It provides diversity to model $y_i=C_i(kt)$, such as trigonometric functions, polynomial functions and so on, These models are a three-layer fully connected network trained with L2 loss.Each dataset has 10000 training samples, the training period is 100, and the batch size in the experiment is set to 40.

The second step is to train the TSDFNet using the ADAM\cite{b40} optimizer with an initial learning rate of 0.0001. The dropout rate \cite{b57} is set to 0.1 for better generalization, and batch size is set to 32. Early stopping is employed to avoid overfitting, the training process will be terminated if no loss degradation within 10 epochs. Metrics were RAE and SMAPE defined by

\begin{flalign}
 RAE = \frac{{\sum\nolimits_{(i,t) \in \Omega Test} {|Y_{it} - } \hat Y_{it}|}}{{\sum\nolimits_{(i,t) \in \Omega Test} {|Y_{it} - } mean({\bf{Y}})|}} 
\end{flalign}
\begin{flalign}
 SMAPE = \frac{{2\sum\nolimits_{(i,t) \in \Omega Test} {|Y_{it} - \hat Y_{it}} |}}{{N\sum\nolimits_{(i,t) \in \Omega Test} {|Y_{it}| + |\hat Y_{it}} |}} 
\end{flalign}

\subsection{Univariate time series prediction and result analysis }

We list the univariate results of five typical datasets in Table \ref{table:1} and Table \ref{table:2}. The peaks and valleys of uncertainty are difficult to forecast since the electrocardiogram and energy consumption are quasi-periodic. The difficulty is that there are so few training samples that cycles and trends in the data on retail sales and air passenger numbers cannot be reliably predicted. On the other hand,although sunspot eruptions are periodic, the extremely long time horizon makes predictions impossible.

\begin{figure}[htbp]
 \centering
 \includegraphics[width=\linewidth]{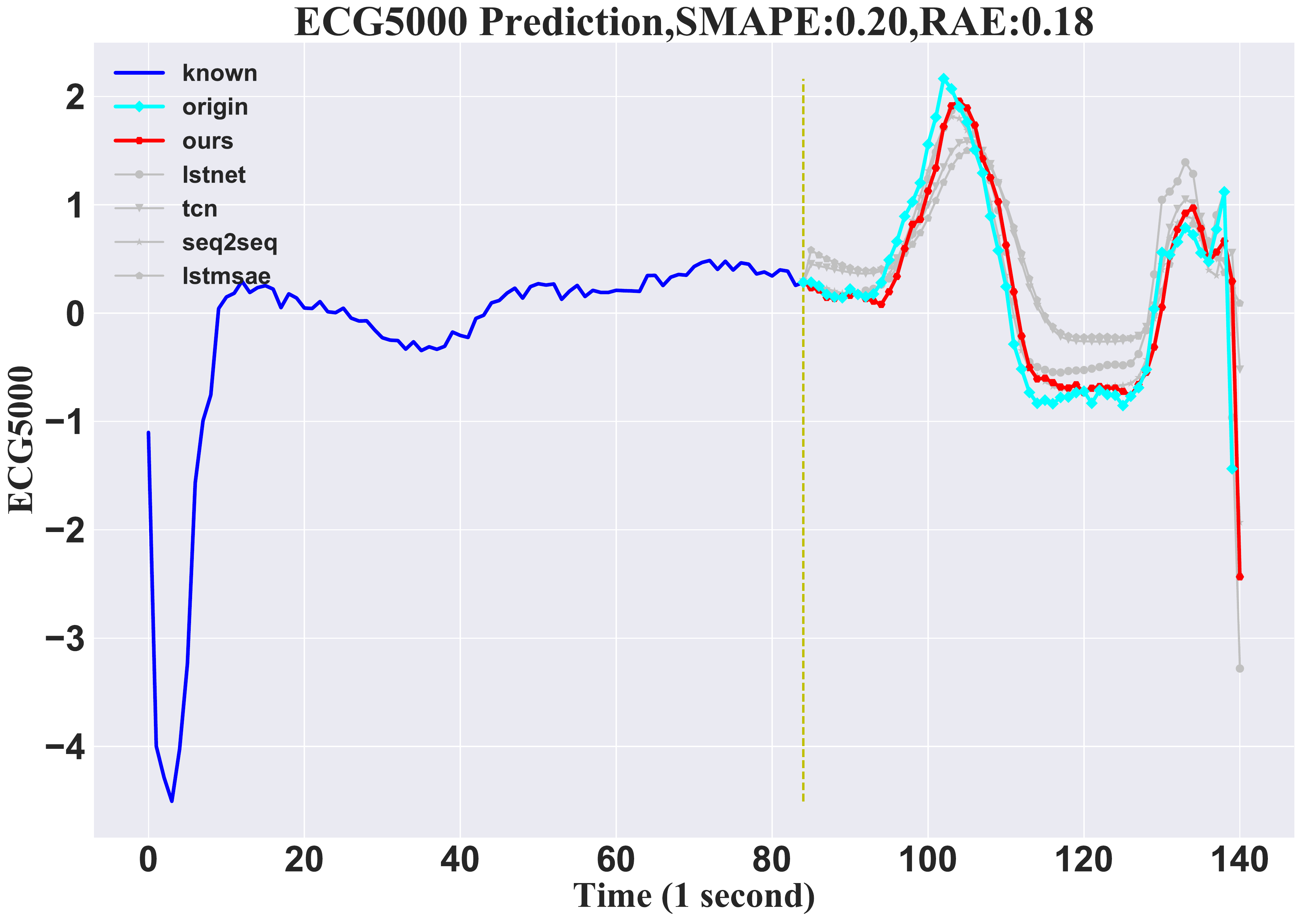}
 \caption{Forecast result of the ECG5000}
 \label{fig:4-1}
\end{figure}

\textbf{ECG5000 } A total of 5000 electrocardiograms (ECGs) ,4500 samples for training and 500 samples for testing, make up the dataset, which is derived from UCR time series. Each sample in the sequence is 140 time points. The experiment uses the first 84 time steps as input, while the final 56 time steps are predicted as the result. The result is shown in Figure \ref{fig:4-1}.

\begin{figure}[htbp]
 \centering
 \includegraphics[width=\linewidth]{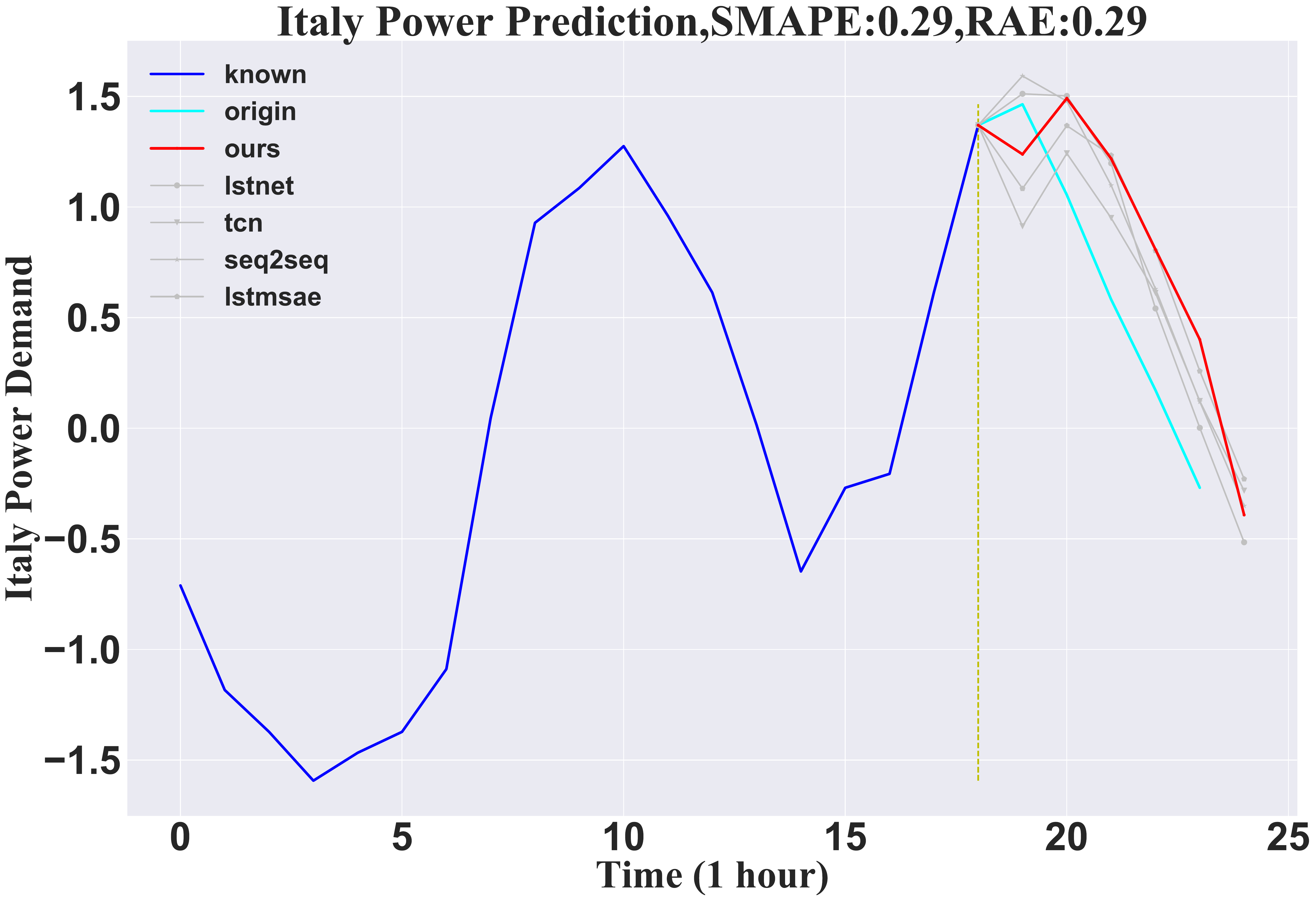}
 \caption{ItalyPowerDemand forecast result}
 \label{fig:4-2}
\end{figure}

\textbf{ItalyPowerDemand }Dataset is derived from a 12-month time series of Italy's power demand. 67 samples were evaluated after 1029 samples had been trained. There are 24 samples total in each set. In the experiment, the data from the first 18 hours are utilized as an input to forecast the data from the following 6 hours. The result is shown in Figure \ref{fig:4-2}.

\begin{figure}[htbp]
 \centering
 \includegraphics[width=\linewidth]{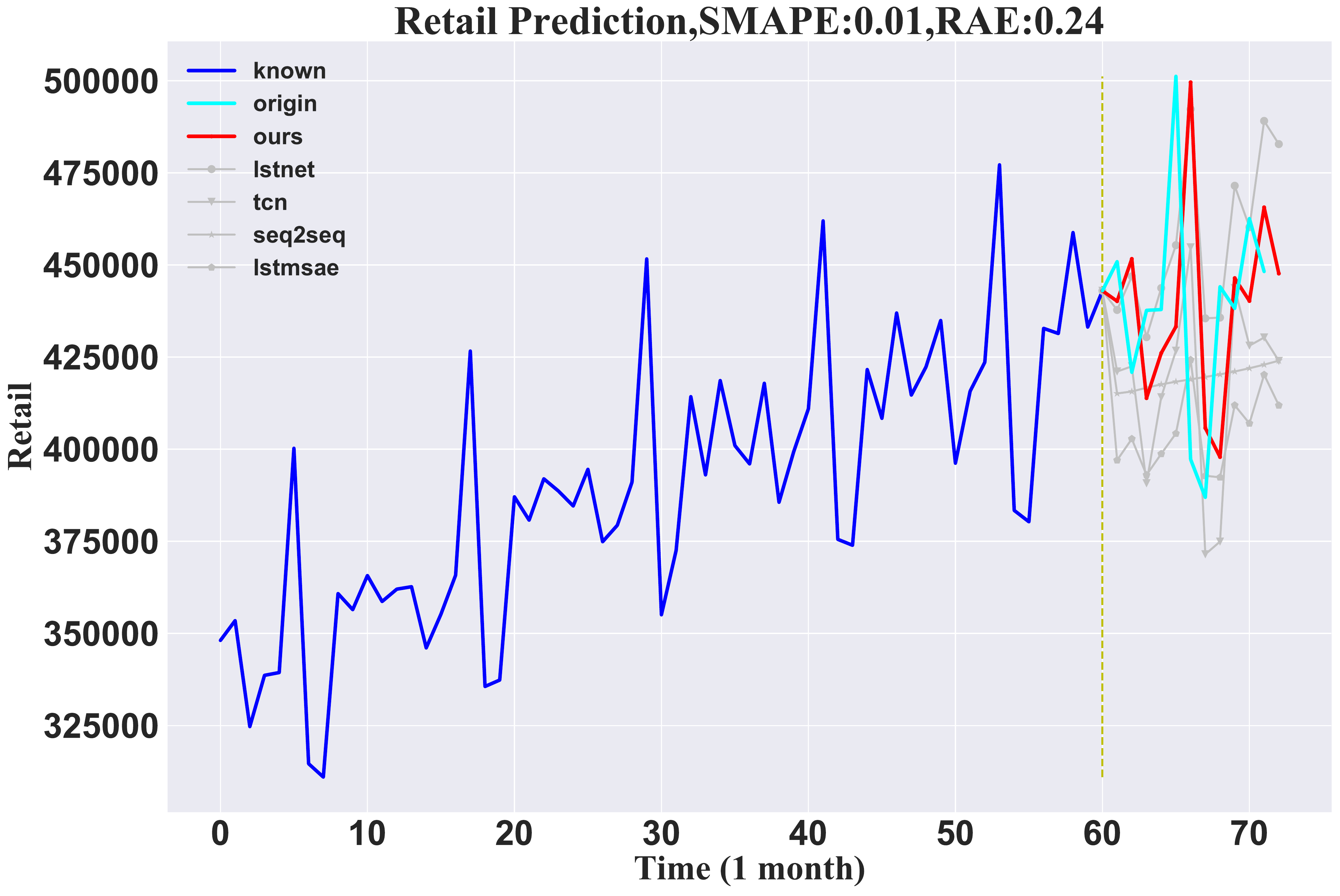}
 \caption{Forecast results of Retail}
 \label{fig:4-3}
\end{figure}

\textbf{Retail }Dataset from Kaggle provides US monthly retail sales data from January 1, 1992 to May 1, 2016. It has a total of 293 samples. 95 percent of the data were used as the training set and 5 percent as the test set in this paper. In order to forecast the data for the following six months, the data from the preceding 12 months were used as input. The result is shown in Figure \ref{fig:4-3}.

\begin{figure}[htbp]
 \centering
 \includegraphics[width=\linewidth]{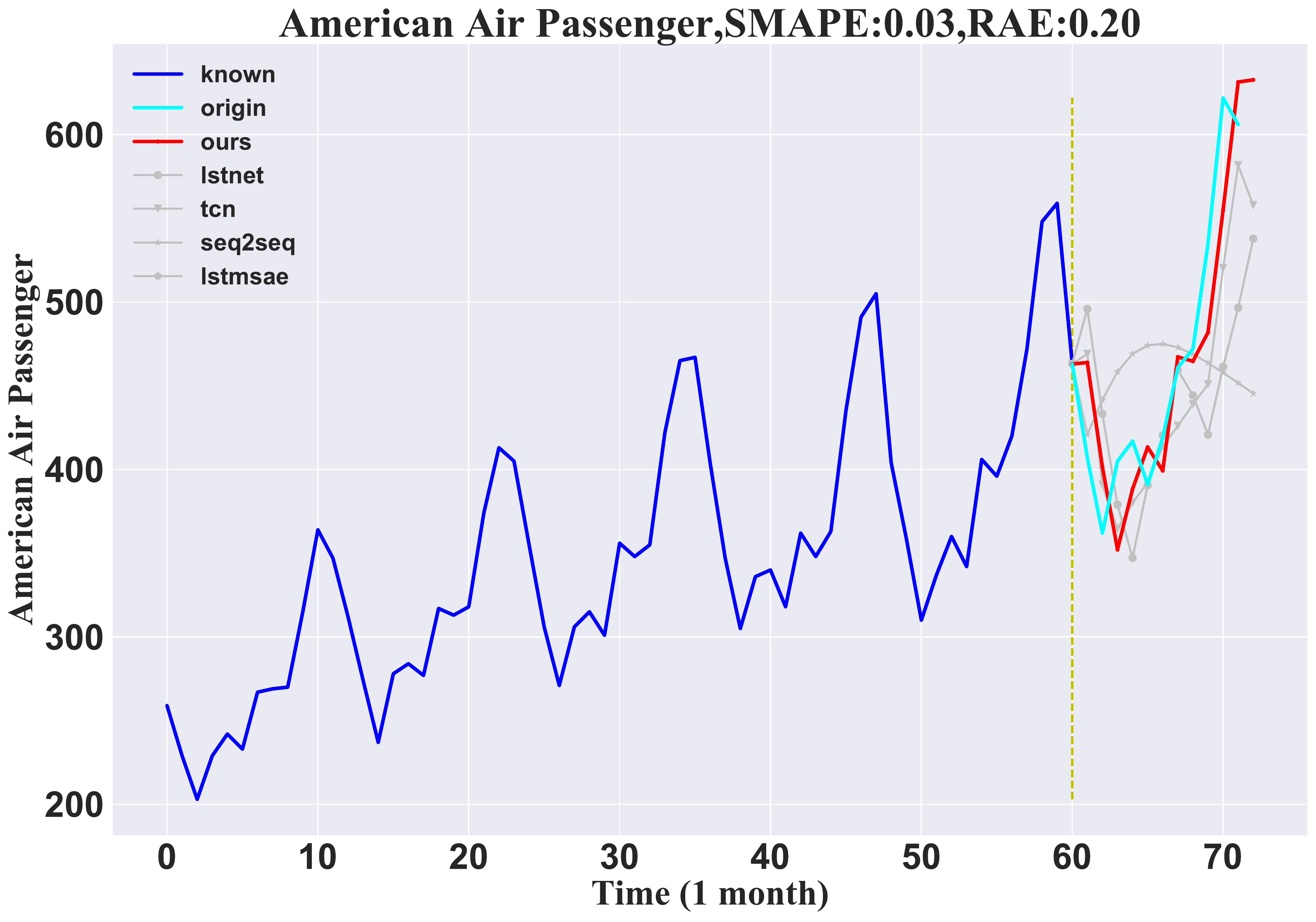}
 \caption{Forecast results of Airpassenger}
 \label{fig:4-4}
\end{figure}

\textbf{Airpassenger }The dataset provides monthly passenger counts for American Airlines on foreign airlines from 1949 to 1960. There are 144 samples in all. The training set consisted of the data from 1949 to 1959, while the test set comprised the data from 1960. The input data spans 60 months, and a 12-month projection is made for the data. The result is shown in Figure \ref{fig:4-4}.

\begin{figure}[htbp]
 \centering
 \includegraphics[width=\linewidth]{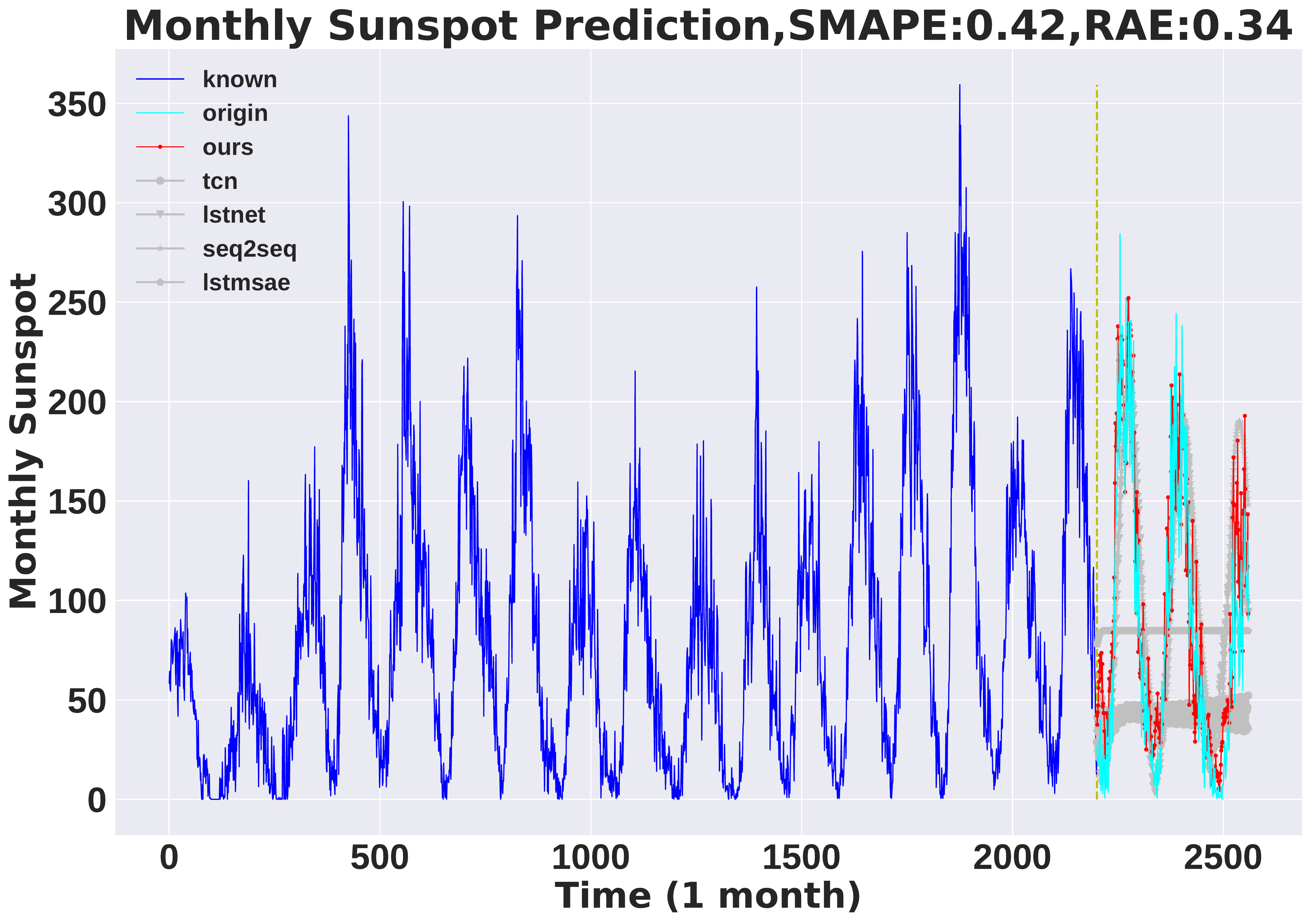}
 \caption{Forecast results of Sunspot}
 \label{fig:4-5}
\end{figure}

\textbf{Sunspot }Dataset is comprised of monthly sunspot observations
(1749-2019) with 2820 samples. 95\% were used as the training set and the last 5\% as the test set. The length of input data is 2200 months, and 360 months of future data are predicted. The result is shown in Figure \ref{fig:4-5}.

\begin{table}[htbp]
\captionsetup{font=normalsize}
\normalsize
\centering

\caption{ RAE of univariate time series models}

\vspace{3pt} 
\begin{tabular}{p{45pt}p{28pt}p{28pt}p{28pt}p{28pt}p{28pt}}
 \hline
 \parbox{45pt}{} & \parbox{28pt}{
 Seq2Seq
 } & \parbox{28pt}{
 LstNet
 } & \parbox{28pt}{
 TCN
 } & \parbox{28pt}{
 SAE
 } & \parbox{28pt}{
 Ours
 } \\
 \hline
 \parbox{45pt}{
 ECG5000
 } & \parbox{28pt}{
 0.360
 } & \parbox{28pt}{
 0.669
 } & \parbox{28pt}{
 0.793
 } & \parbox{28pt}{
 0.938
 } & \parbox{28pt}{
 \textbf{0.290}
 } \\
 \parbox{45pt}{
 Power
 } & \parbox{28pt}{
 0.294
 } & \parbox{28pt}{
 0.276
 } & \parbox{28pt}{
 0.718
 } & \parbox{28pt}{
 0.425
 } & \parbox{28pt}{
 \textbf{0.162}
 } \\
 \parbox{45pt}{
 Retail
 } & \parbox{28pt}{
 1.803
 } & \parbox{28pt}{
 1.536
 } & \parbox{28pt}{
 0.855
 } & \parbox{28pt}{
 1.426
 } & \parbox{28pt}{
 \textbf{0.201}
 } \\
 \parbox{45pt}{
 AirPassenger
 } & \parbox{28pt}{
 1.233
 } & \parbox{28pt}{
 0.993
 } & \parbox{28pt}{
 1.911
 } & \parbox{28pt}{
 1.342
 } & \parbox{28pt}{
 \textbf{0.252}
 } \\
 \parbox{45pt}{
 Sunspot
 } & \parbox{28pt}{
 0.644
 } & \parbox{28pt}{
 0.904 
 } & \parbox{28pt}{
 0.671
 } & \parbox{28pt}{
 0.743
 } & \parbox{28pt}{
 \textbf{0.334}
 } \\
 \hline
 \label{table:1}
\end{tabular}
\vspace{2pt}
\end{table}

\begin{table}[htbp]
 \captionsetup{font=normalsize}
 \normalsize
 \centering
 \caption{SMAPE of univariate time series models}
\vspace{3pt} 
\begin{tabular}{p{45pt}p{28pt}p{28pt}p{28pt}p{28pt}p{28pt}}
 \hline
 \parbox{45pt}{} & \parbox{28pt}{
 Seq2Seq
 } & \parbox{28pt}{
 LstNet
 } & \parbox{28pt}{
 TCN
 } & \parbox{28pt}{
 SAE
 } & \parbox{28pt}{
 Ours
 } \\
 \hline
 \parbox{45pt}{
 ECG5000 
 } & \parbox{28pt}{
 0.424
 } & \parbox{28pt}{
 0.685
 } & \parbox{28pt}{
 0.682
 } & \parbox{28pt}{
 0.731
 } & \parbox{28pt}{
 \textbf{0.373}
 } \\
 \parbox{45pt}{
 Power
 } & \parbox{28pt}{
 0.334
 } & \parbox{28pt}{
 0.364
 } & \parbox{28pt}{
 0.606
 } & \parbox{28pt}{
 0.375
 } & \parbox{28pt}{
 \textbf{0.144}
 } \\
 \parbox{45pt}{
 Retail 
 } & \parbox{28pt}{
 0.052
 } & \parbox{28pt}{
 0.115
 } & \parbox{28pt}{
 0.098
 } & \parbox{28pt}{
 0.158
 } & \parbox{28pt}{
 \textbf{0.013}
 } \\
 \parbox{45pt}{
 AirPassenger
 } & \parbox{28pt}{
 0.132 
 } & \parbox{28pt}{
 0.086 
 } & \parbox{28pt}{
 0.213
 } & \parbox{28pt}{
 0.282
 } & \parbox{28pt}{
 \textbf{0.027}
 } \\
 \parbox{45pt}{
 Sunspot
 } & \parbox{28pt}{
 0.577 
 } & \parbox{28pt}{
 0.847
 } & \parbox{28pt}{
 0.573 
 } & \parbox{28pt}{
 0.757
 } & \parbox{28pt}{
 \textbf{0.431}
 } \\
 \hline
 \label{table:2}
\end{tabular}
\vspace{2pt}
\end{table}

\subsection{Multivariate time series prediction and result analysis}
We list the multivariable results of five typical datasets in Table \ref{table:3} and Table \ref{table:4}.
Weather datasets, traffic datasets, wind speed datasets derived from actual observations, and two artificial Lorenz datasets are examples of multivariable datasets. These multivariable data are often typical chaotic systems, which are more complex than univariate data, necessitating the addition of additional features to help with prediction.
In order to more effectively demonstrate the issue, Lorenz flalign created a Lorenz data set where relevant parameters could be statistically adjusted. The target variable in the weather data set, temperature, has various modal periodicities and is influenced by numerous additional factors, making it a good choice to demonstrate the model's interpretability. Due to the very high degree of uncertainty in the traffic data set and the wind speed data set, future known factors, such as wind speed and average vehicle speed from other observation stations, must be employed to aid prediction. While other models also include future known elements, the results are subpar, demonstrating that the model in this work has a strong understanding of spatial information features.

\begin{figure}[htbp]
 \centering
 \includegraphics[width=\linewidth]{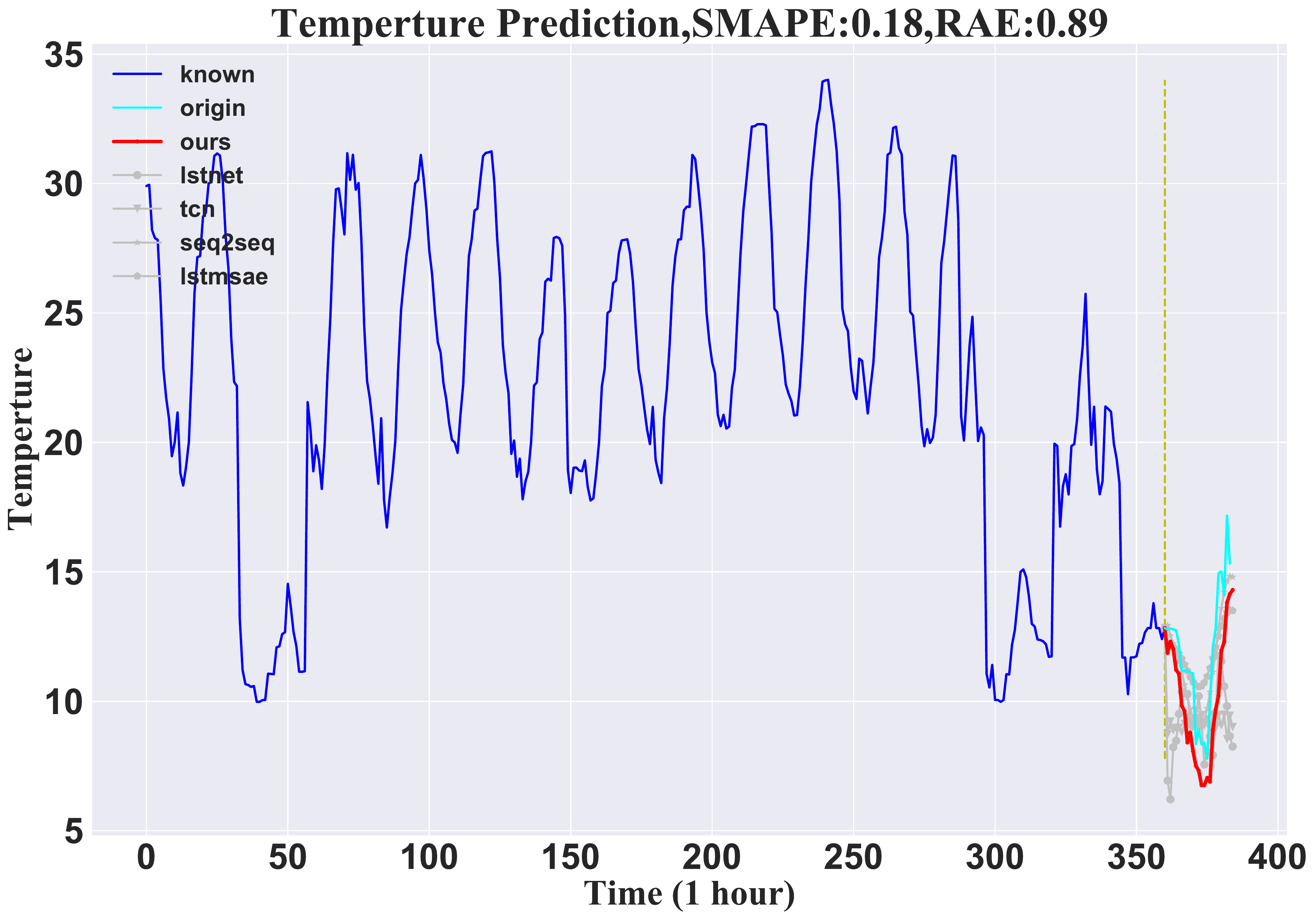}
 \caption{Forecast results of weather}
 \label{fig:4-10}
\end{figure}

\textbf{Weather} Dataset, which spans the years 2006 to 2016, was obtained from Kaggle. The experiment's temperature prediction is dependent on other variables such as humidity, wind speed, wind direction, visibility, cloud cover, air pressure, etc. A total of 96453 samples were included in the data, which were split up into hours. 95 percent of the data were utilized for training, and 5 percent were used for testing. The model predicted the temperature for the following 360 hours using data from the first 2,200 hours and the following 360 hours on external features. The result is shown in Figure \ref{fig:4-10}.

\begin{figure}[htbp]
 \centering
 \includegraphics[width=\linewidth]{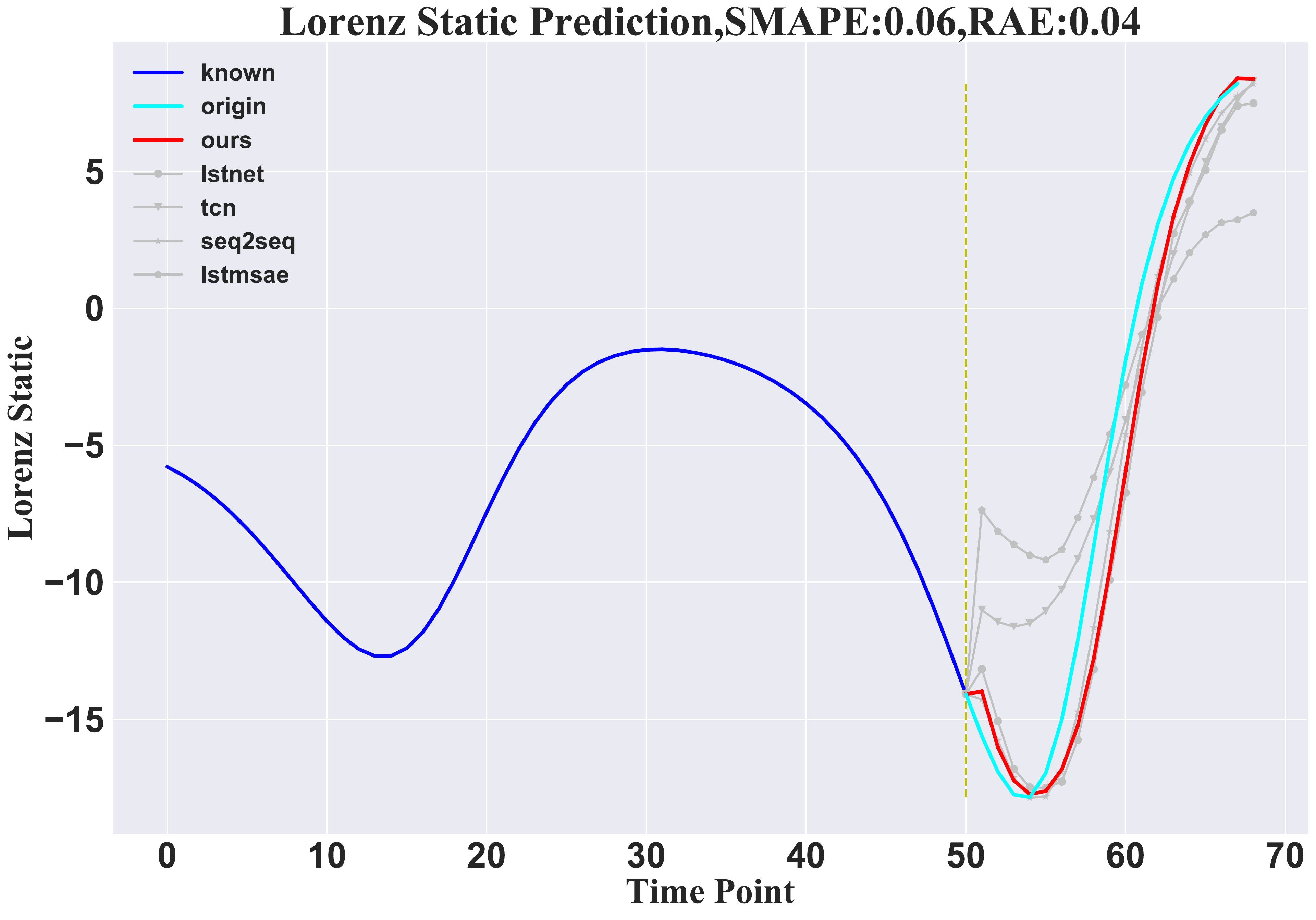}
 \caption{Forecast results of time invariant lorenz system}
 \label{fig:4-11}
\end{figure}

\begin{figure}[htbp]
 \centering
 \includegraphics[width=\linewidth]{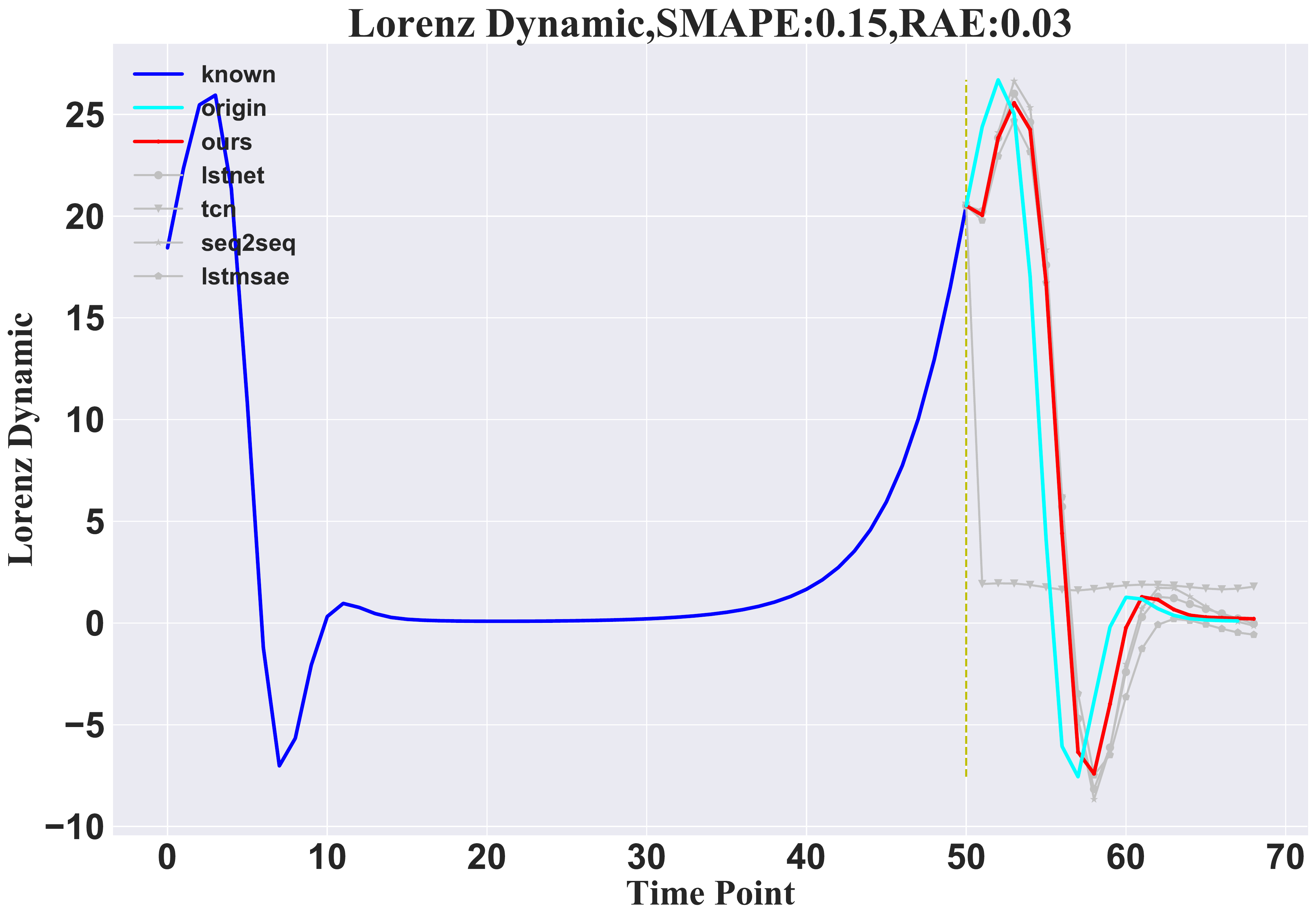}
 \caption{Forecast results of time variant lorenz system}
 \label{fig:4-12}
\end{figure}

\textbf{Lorenz} A 90-dimensional coupled Lorenz model $\mathbf{X}'(t) = G(\mathbf{X}(t); \mathbf{P})$ is used to generate the synthesis time data set under different noise conditions,where $G(\cdot)$ is the nonlinear function set of the Lorenz system with $\mathbf{X}(t)=(x^t_1,...,x^t_{90})'$, $\mathbf{P}$ is the parameter vector. To demonstrate the distinction between the model described in this work in the time-varying system and the time-invariant system, the time-invariant Lorentz system and time-varying Lorentz system are tested in this article, respectively. For a time-invariant Lorenz system (Lorenz-S), $\mathbf{P}$ does not change over time; however, for time-varying systems (Lorenz-D), $\mathbf{P}$ does. In the experiment simulation, 5000 samples were generated; the first 90\% of the data were utilized as the training set, and the final 10\% as the test set. The experimental findings demonstrate that the model presented in this paper functions effectively in both time-varying and time-invariant systems. The result is shown in Figure \ref{fig:4-11} and Figure \ref{fig:4-12}.

\begin{figure}[htbp]
 \centering
 \includegraphics[width=\linewidth]{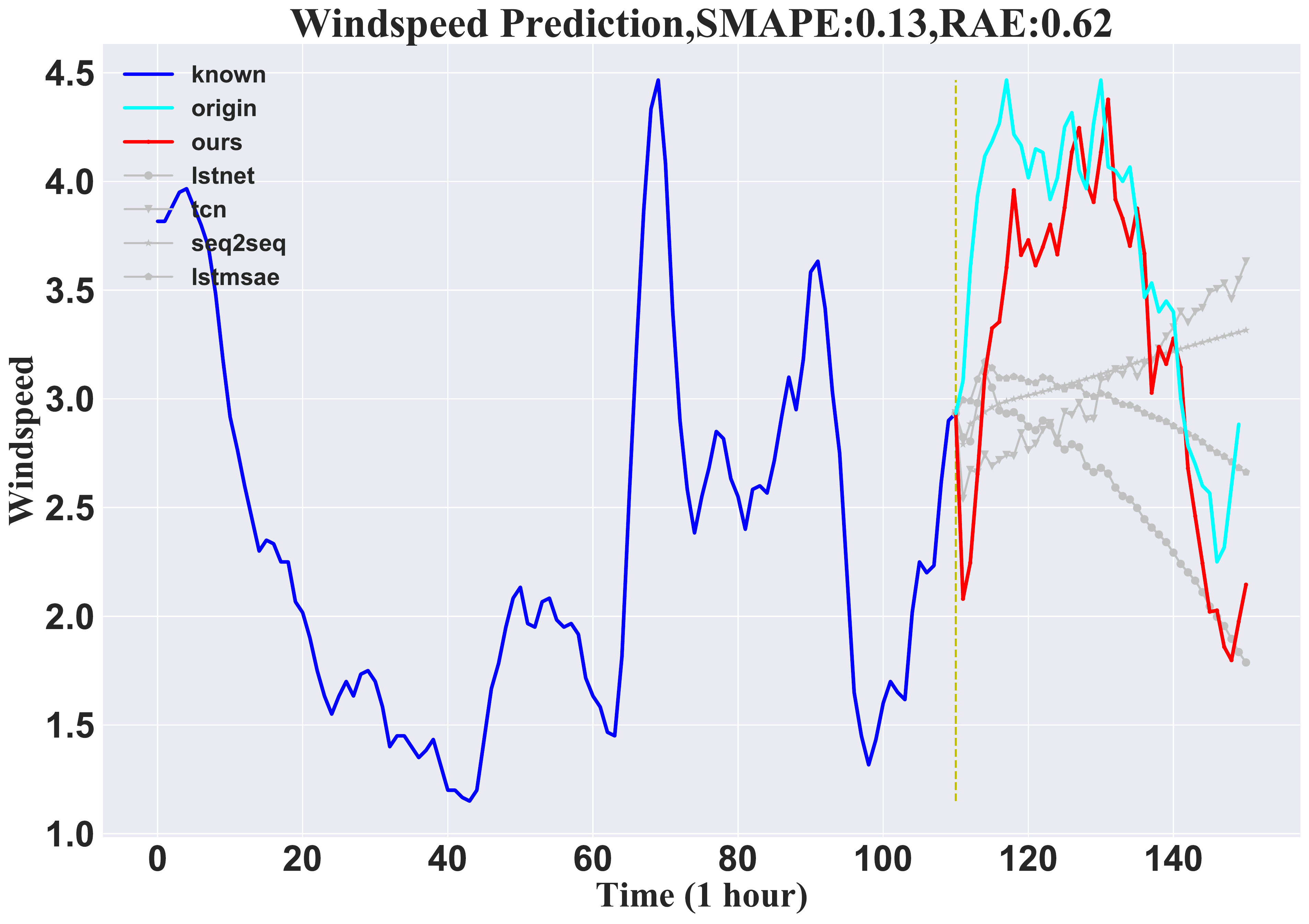}
 \caption{Forecast results of windspeed}
 \label{fig:4-13}
\end{figure}

\textbf{Windspeed} This dataset, which is a high dimensional (155-dimensional) windspeed data set from 155 stations in Wakkanai, Japan, is supplied by the Japan Meteorological Agency. Every 10 minutes, for a total of 138,600 minutes, are recorded. The known duration of 110 minutes is used in this paper to estimate the wind speed for the subsequent 40 minutes. Data from 154 other sites were used as additional features, and a target wind speed at one of 155 sites was chosen at random. 95\% of the initial data were utilized as a training set,  while 5\% were used as a test set.Even though wind speed is typically thought to be exceedingly challenging to predict. This approach predicts outcomes that are superior to those anticipated by other methods. The result is shown in Figure \ref{fig:4-13}.

\begin{figure}[htbp]
 \centering
 \includegraphics[width=\linewidth]{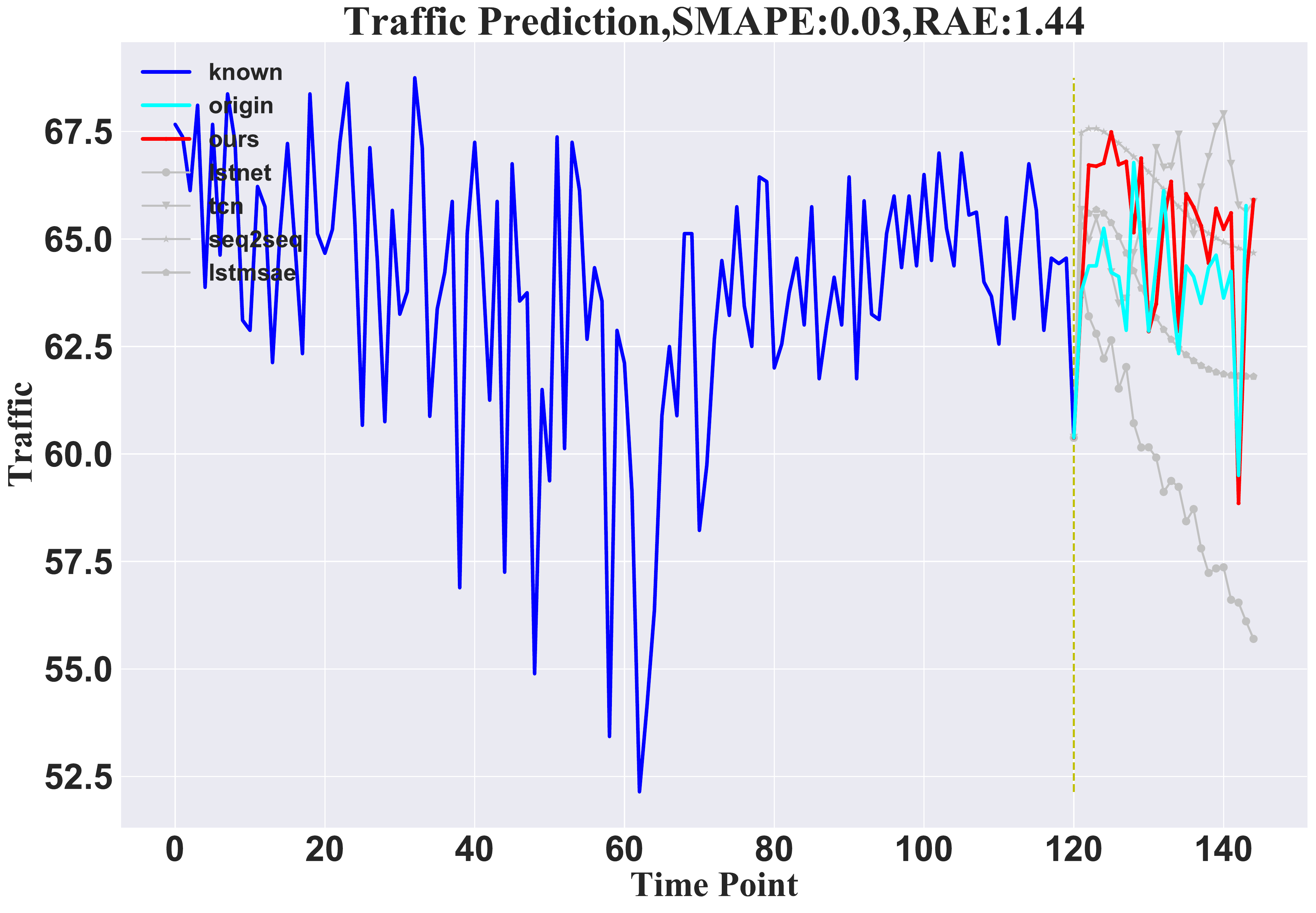}
 \caption{Forecast results of traffic}
 \label{fig:4-14}
\end{figure}

\textbf{Traffic }The average vehicle speed (MPH) is predicted using data from 207 loop detectors along Route 134 in Los Angeles, with the observations from each detector being handled as a separate variable. The first 120 hours are utilized to forecast the 24-hour average speed. From the high-dimensional data, the data from one sensor is chosen as the target variable, while the observations from other sensors are used as supplementary features. It demonstrates the model's capacity to forecast spatial data. The result is shown in Figure \ref{fig:4-14}.

\begin{center}
 
\end{center}

\begin{table}[htbp]
 \captionsetup{font=normalsize}
 \normalsize
 \centering
 \caption{RAE of multivariate time series models}
 
\vspace{3pt} 
\begin{tabular}{p{45pt}p{28pt}p{28pt}p{28pt}p{28pt}p{28pt}}
 \hline
 \parbox{45pt}{} & \parbox{28pt}{
 Seq2Seq
 } & \parbox{28pt}{
 Lstnet
 } & \parbox{28pt}{
 TCN
 } & \parbox{28pt}{
 SAE
 } & \parbox{28pt}{
 Ours
 } \\
 \hline
 \parbox{45pt}{
 Lorenz-S
 } & \parbox{28pt}{
 0.202
 } & \parbox{28pt}{
 0.109
 } & \parbox{28pt}{
 0.474 
 } & \parbox{28pt}{
 0.685
 } & \parbox{28pt}{
 \textbf{0.060}
 } \\
 \parbox{45pt}{
 Lorenz-D
 } & \parbox{28pt}{
 0.888
 } & \parbox{28pt}{
 1.174
 } & \parbox{28pt}{
 1.574
 } & \parbox{28pt}{
 1.230
 } & \parbox{28pt}{
 \textbf{0.624}
 } \\
 \parbox{45pt}{
 Weather
 } & \parbox{28pt}{
 0.305
 } & \parbox{28pt}{
 1.985
 } & \parbox{28pt}{
 1.822
 } & \parbox{28pt}{
 1.204
 } & \parbox{28pt}{
 \textbf{0.218}
 } \\
 \parbox{45pt}{
 WindSpeed
 } & \parbox{28pt}{
 0.294
 } & \parbox{28pt}{
 0.428
 } & \parbox{28pt}{
 0.531
 } & \parbox{28pt}{
 0.284
 } & \parbox{28pt}{
 \textbf{0.094}
 } \\
 \parbox{45pt}{
 Traffic
 } & \parbox{28pt}{
 0.262
 } & \parbox{28pt}{
 0.241
 } & \parbox{28pt}{
 0.241
 } & \parbox{28pt}{
 0.588
 } & \parbox{28pt}{
 \textbf{0.074}
 } \\
 \hline
 \label{table:3}
\end{tabular}
\vspace{2pt}
\end{table}

\begin{table}[htbp]
 \captionsetup{font=normalsize}
 \normalsize
 \centering
 \caption{SMAPE of multivariable time series models}

\vspace{3pt} 
\begin{tabular}{p{45pt}p{28pt}p{28pt}p{28pt}p{28pt}p{28pt}}
 \hline
 \parbox{45pt}{} & \parbox{28pt}{
 Seq2Seq
 } & \parbox{28pt}{
 Lstnet
 } & \parbox{28pt}{
 TCN
 } & \parbox{28pt}{
 SAE
 } & \parbox{28pt}{
 Ours
 } \\
 \hline
 \parbox{45pt}{
 Lorenz-S
 } & \parbox{28pt}{
 0.448
 } & \parbox{28pt}{
 0.216
 } & \parbox{28pt}{
 0.517
 } & \parbox{28pt}{
 0.487
 } & \parbox{28pt}{
 \textbf{0.101}
 } \\
 \parbox{45pt}{
 Lorenz-D
 } & \parbox{28pt}{
 0.785
 } & \parbox{28pt}{
 1.292
 } & \parbox{28pt}{
 2.675
 } & \parbox{28pt}{
 1.080
 } & \parbox{28pt}{
 \textbf{0.675}
 } \\
 \parbox{45pt}{
 Weather
 } & \parbox{28pt}{
 5.151
 } & \parbox{28pt}{
 7.338
 } & \parbox{28pt}{
 4.565
 } & \parbox{28pt}{
 4.426
 } & \parbox{28pt}{
 \textbf{0.899}
 } \\
 \parbox{45pt}{
 Windspeed
 } & \parbox{28pt}{
 3.254
 } & \parbox{28pt}{
 5.042
 } & \parbox{28pt}{
 6.442
 } & \parbox{28pt}{
 1.468
 } & \parbox{28pt}{
 \textbf{0.624}
 } \\
 \parbox{45pt}{
 Traffic
 } & \parbox{28pt}{
 5.156
 } & \parbox{28pt}{
 5.338
 } & \parbox{28pt}{
 5.189
 } & \parbox{28pt}{
 5.741
 } & \parbox{28pt}{
 \textbf{0.899}
 } \\
 \hline
 \label{table:4}
\end{tabular}
\vspace{2pt}
\end{table}

For better comparison, we follow the experimental results of Automatic Reserve Pool
Neural Network (ARNN)\cite{b49} directly instead of reproducing the paper, which has been tested on Lorenz, windspeed and traffic datasets, using RMSE as the metric. The result is shown in Table \ref{table:5}.

\begin{table}[htbp]
 \captionsetup{font=normalsize}
 \normalsize
 \centering
 \caption{ARNN comparison of multivariate time series}
 
\vspace{3pt} 
\begin{tabular}{p{45pt}p{40pt}p{40pt}p{40pt}p{25pt}}
 \hline
 \parbox{45pt}{} & \parbox{40pt}{
 Lorenz-S
 } & \parbox{40pt}{
 Lorenz-D
 } & \parbox{40pt}{
 Windspeed
 } & \parbox{25pt}{
 Traffic
 } \\
 \hline
 \parbox{45pt}{
 ARNN
 } & \parbox{40pt}{
 0.397
 } & \parbox{40pt}{
 0.513
 } & \parbox{40pt}{
 0.425
 } & \parbox{25pt}{
 1.01 
 } \\
 \parbox{45pt}{
 Ours
 } & \parbox{40pt}{
 \textbf{0.323 }
 } & \parbox{40pt}{
 \textbf{0.490 }
 } & \parbox{40pt}{
 \textbf{0.375} 
 } & \parbox{25pt}{
 \textbf{1.08}
 } \\
 \hline
 \label{table:5}
\end{tabular}
\vspace{2pt}
\end{table}
 
\begin{figure}[htbp]
 \centering
 \includegraphics[width=\linewidth]{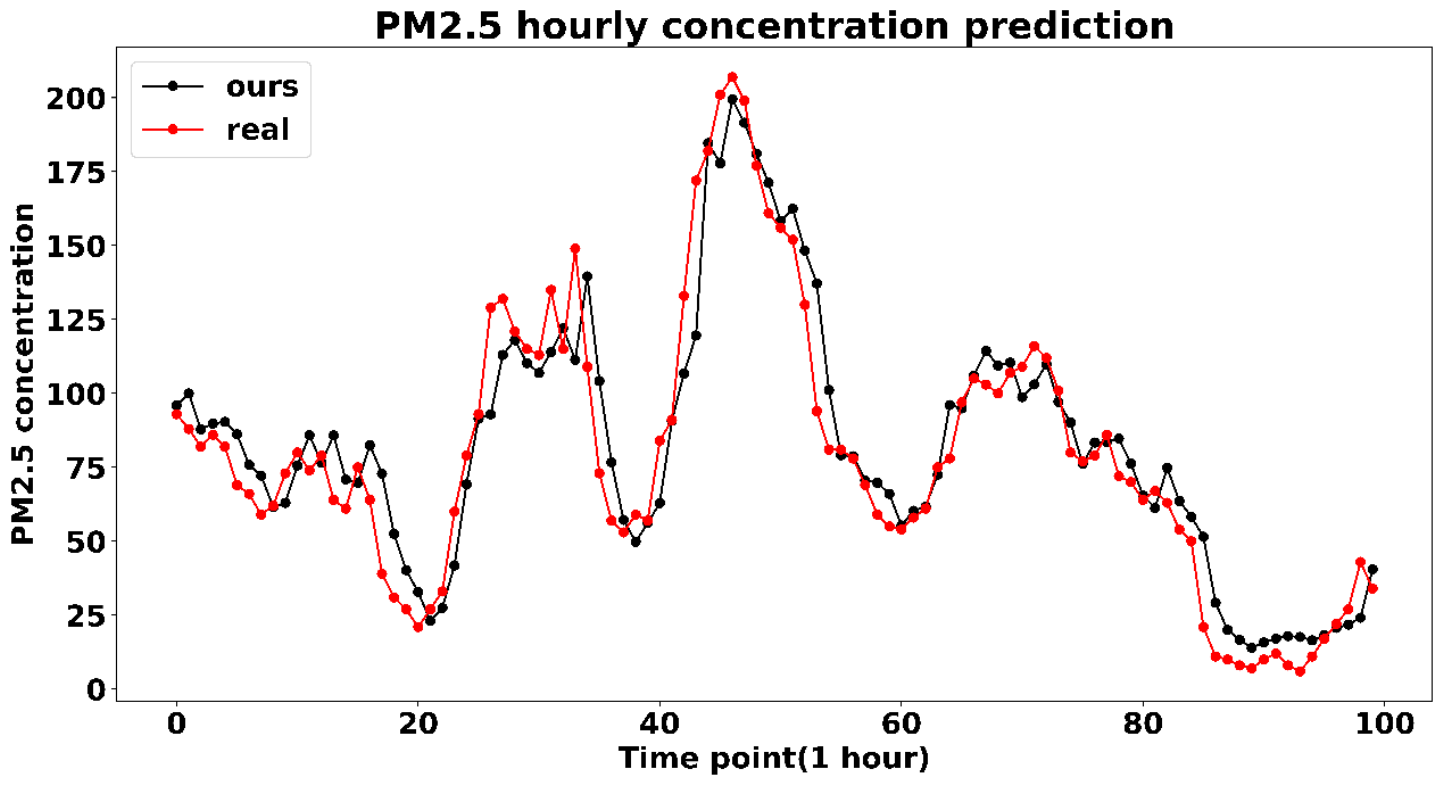}
 \caption{forecast results of PM2.5}
 \label{fig:4-19}
\end{figure}

To show that TSDFNet also performs well on single-step prediction problems, we compare it with LSTM-SAE on air quality datasets. The forecast target is PM2.5 concentration from 2010 to 2014, our model achieves significant RMSE reduction (24→15), The result is shown in Figure \ref{fig:4-19}.

\subsection{Discussion}

After establishing the performance advantages of TSDFNet, we then demonstrate how our model design allows the analysis of its specific componentsto explain the general relationships it has learned.

\begin{figure}[htbp]
 \centering
 \includegraphics[width=\linewidth]{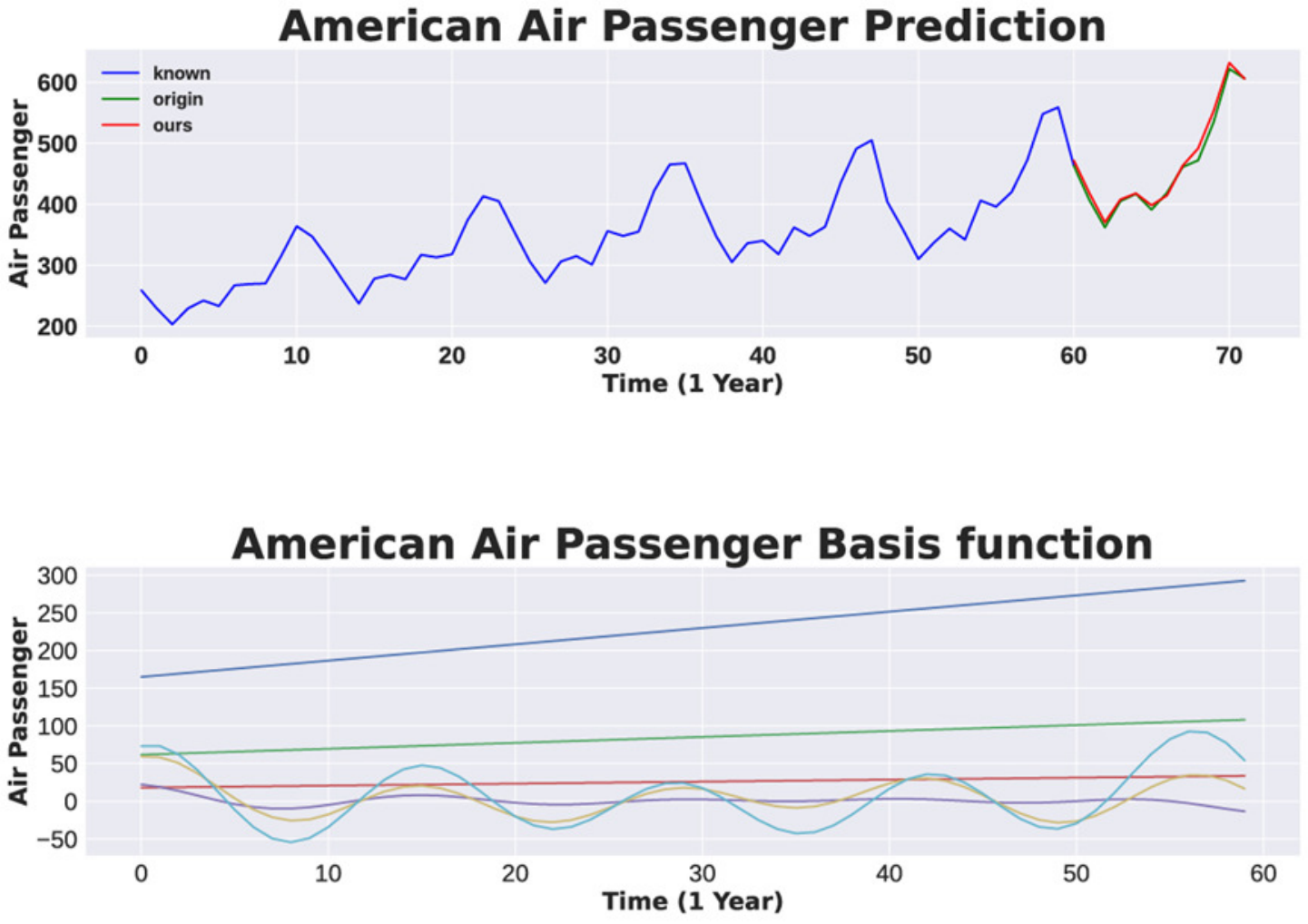}
 \caption{feature analysis of basis function of American air passenger}
 \label{fig:4-8}
\end{figure}

We first quantify feature importance by analyzing the interpretable variants of TDN
described in Eq.(\ref{eq:wn}).
The air passenger dataset is not stationary and combines seasonality and trend. It grows yearly and varies frequently from month to month. The outcome is depicted in Figure \ref{fig:4-8}, The original signal and its prediction are displayed in the figure's top row, while the basis function is displayed in the second row.

\begin{figure}[htbp]
 \centering
 \includegraphics[width=\linewidth]{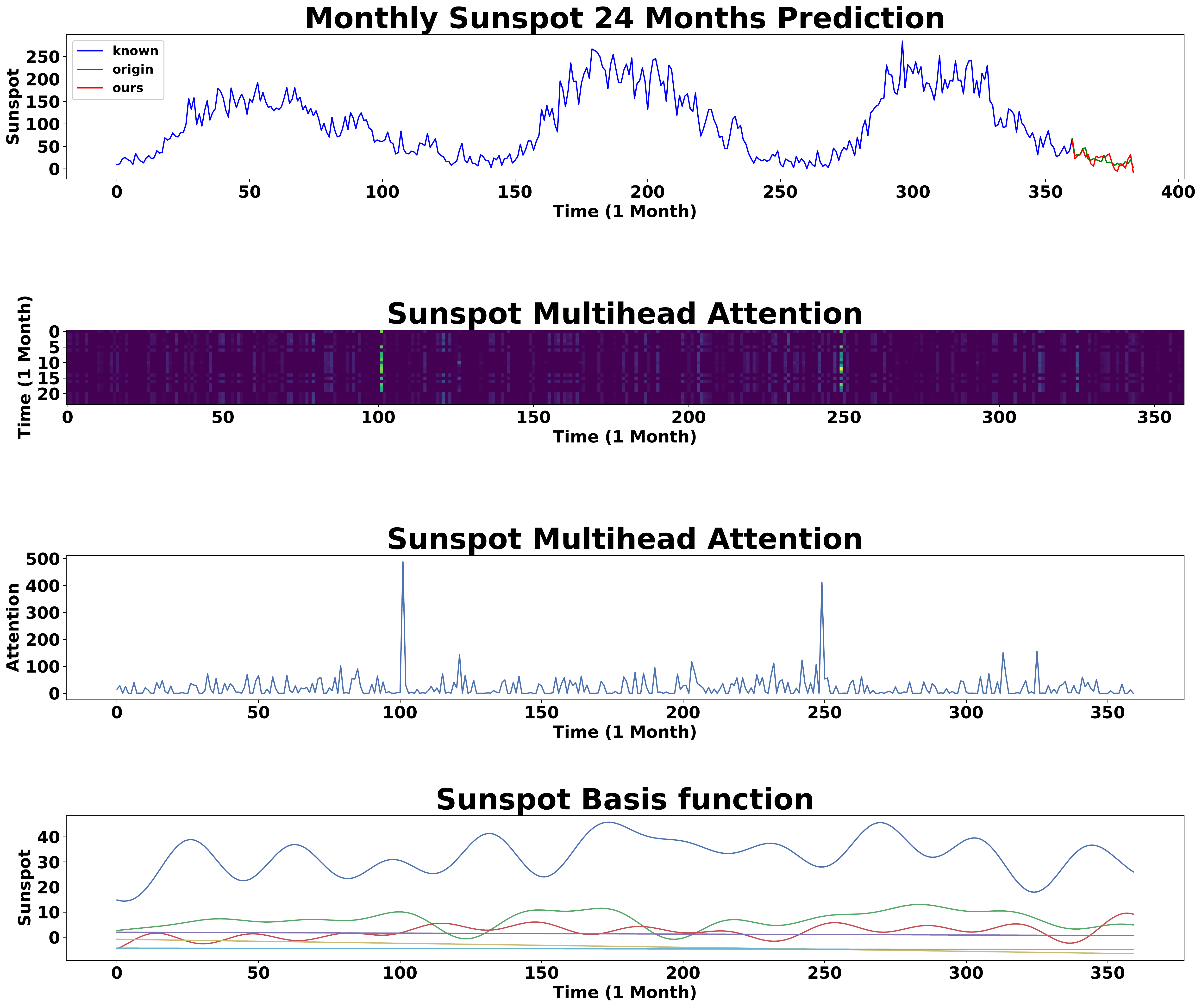}
 \caption{Visualization of sunspot features}
 \label{fig:4-9}
\end{figure}

The outcome for the sunspot dataset is displayed in Figure \ref{fig:4-9}. The top row of the graphic makes evident how seasonal it is, while the bottom row demonstrates how successfully periodicity is captured by the fine-tuned trigonometric basis functions. Additionally, attention weight patterns, which highlight the crucial moments the TSDFNet based its inferences on, might be quite helpful.  We can see many brilliant stripes in the second row, spaced out against a dark background. They stand for the start of a fresh cycle.  It can be seen that the attention spikes match to the troughs of the sunspot when the selection weights for each time step are aggregated, as illustrated in the third row. 

\begin{figure}[htbp]
 \centering
 \includegraphics[width=\linewidth]{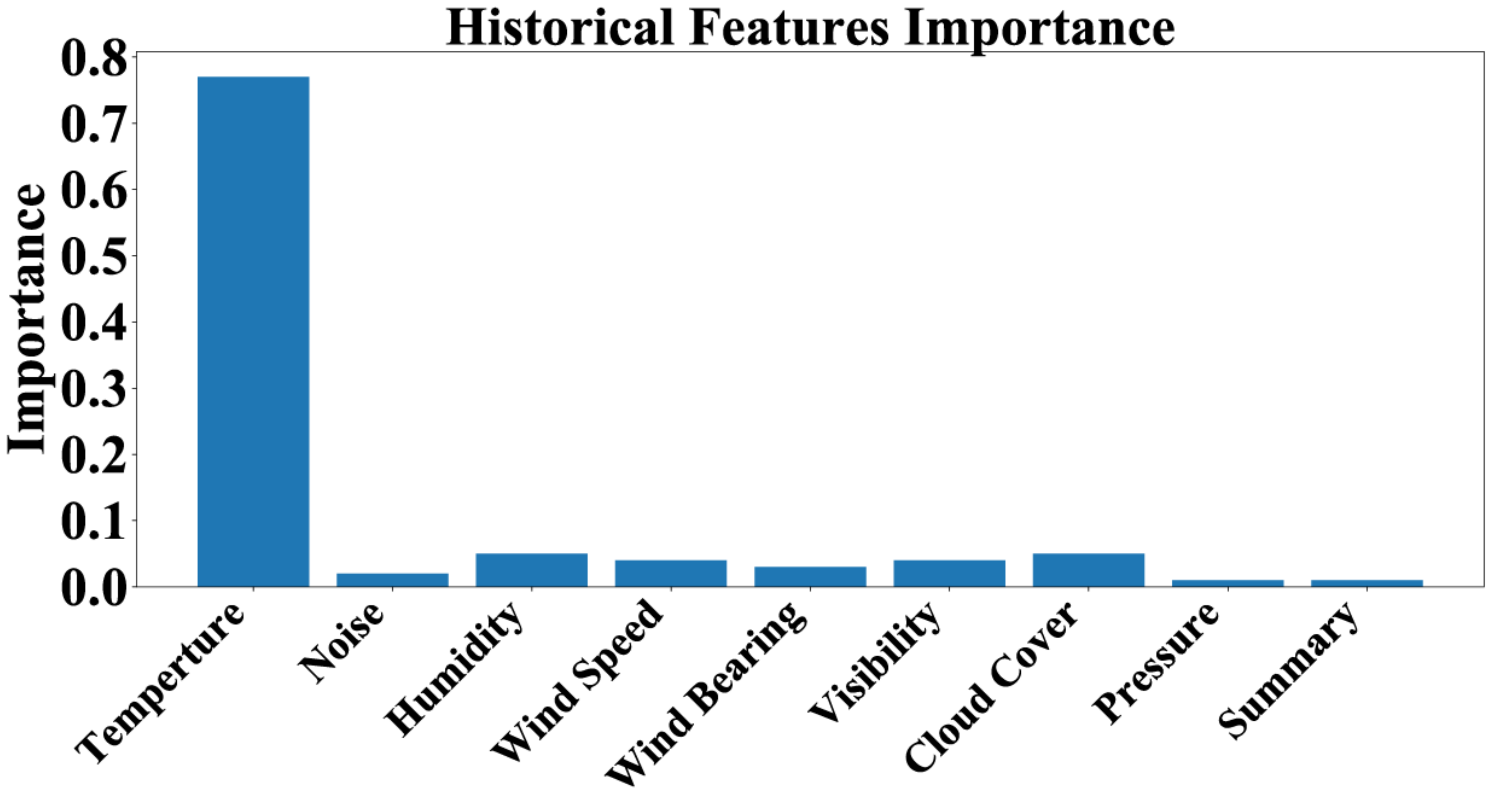}
 \caption{Importance distribution of historical weather data variables}
 \label{fig:4-20}
\end{figure}

\begin{figure}[htbp]
 \centering
 \includegraphics[width=\linewidth]{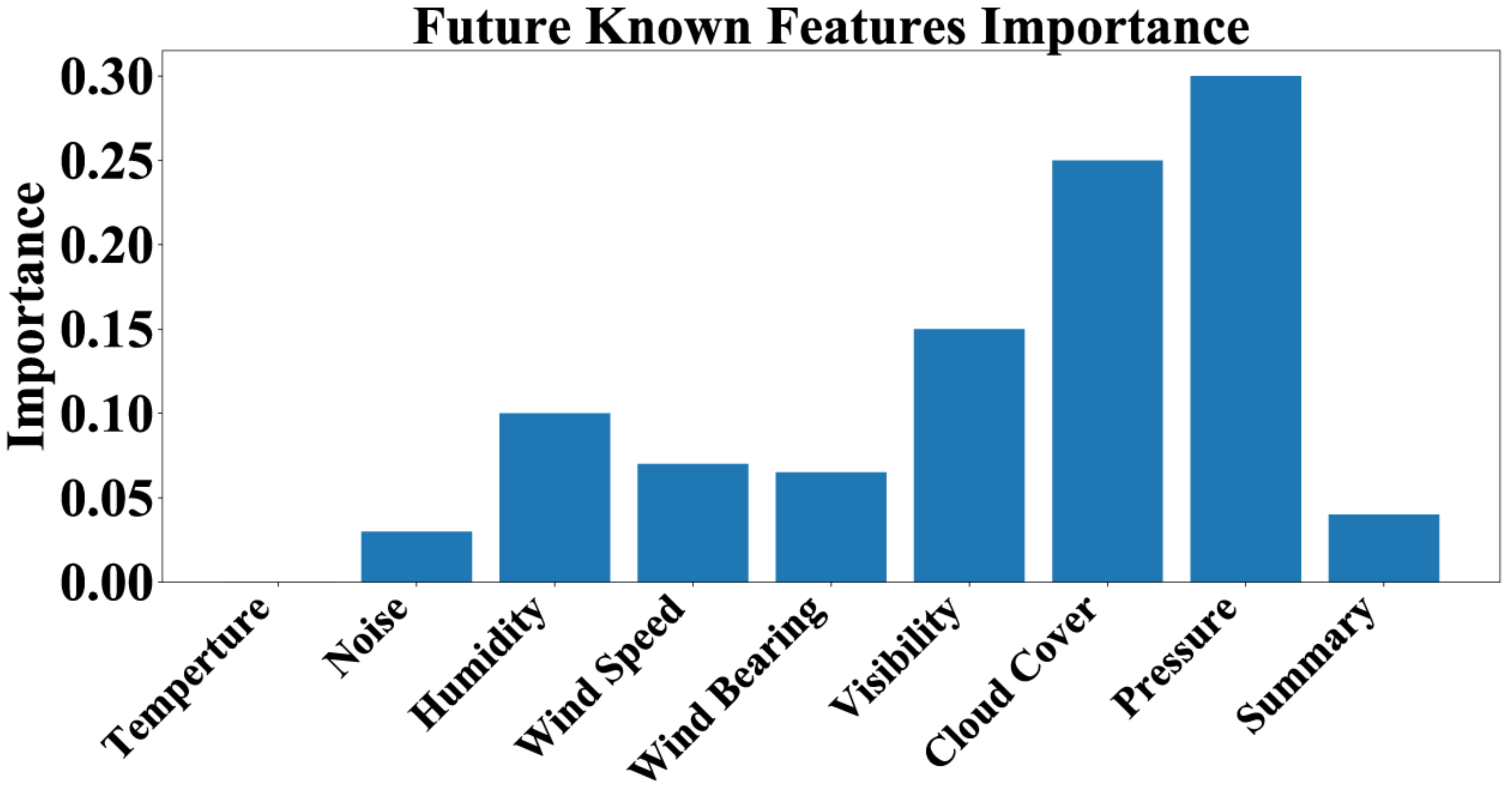}
 \caption{Importance distribution of future weather data related variables}
 \label{fig:4-21}
\end{figure}

Additionally, the importance of each feature, as determined by (\ref{eq:dist}), can be examined using the decision weight patterns in the feature selection block.  The important distribution of historical features is depicted in Figure \ref{fig:4-20} while also accounting for other feature components of historical data.We can see that previous temperature data account for around 75\% of the importance of future temperature data. The important distribution of upcoming known data is depicted in Figure \ref{fig:4-21}.  It is clear that the accuracy of the predictions depends heavily on the air pressure and cloud thickness of the future.  The contribution to the result is also zero because the future temperature data is unknowable and filled with a value of 0.  We also use the Gaussian noise as input to show how our model may exclude unimportant variables.  Gaussian noise's contribution is minimal because it doesn't carry any relevant information. 

\section{Conclusion}

This paper introduces the Temporal Spatial Decomposition Fusion Network (TSDFNet) – a novel interpretable deep learning model which incorporates feature engineering into modeling and achieves incredible performance. TSDFNet utilizes specialized components to handle the long-term forecasting problem of time series, where Temporal Decomposition Network (TDN) allows the user to customize arbitrary basis functions as the eigenmodes of time series decomposition, Spatial Decomposition Network (SDN) uses external features as basis functions for sequence decomposition, Attentive Feature Fusion Network (AFFN) fuses all the input features and selects the most important ones. Finally, extensive experiments demonstrate that the TSDFNet yield consistent state-of-the-art performance in comparison with a variety of well-known algorithms.

\vspace{12pt}

\end{document}